\pdfoutput=1

\documentclass[11pt]{article}

\usepackage[]{EMNLP2023}

\usepackage{times}
\usepackage{latexsym}

\usepackage[T1]{fontenc}

\usepackage[utf8]{inputenc}

\usepackage{microtype}

\usepackage{inconsolata}
\usepackage[greek,english]{babel}
\usepackage{alphabeta}
\usepackage{graphics}
\usepackage{graphicx}
\usepackage{rotating}
\usepackage{multirow}
\usepackage{adjustbox}

\usepackage{array,booktabs}
\usepackage{float}
\usepackage{amsmath}
\usepackage{scalerel,graphicx,xparse}

\NewDocumentCommand\emojicancel{}{{\includegraphics[scale=0.02]{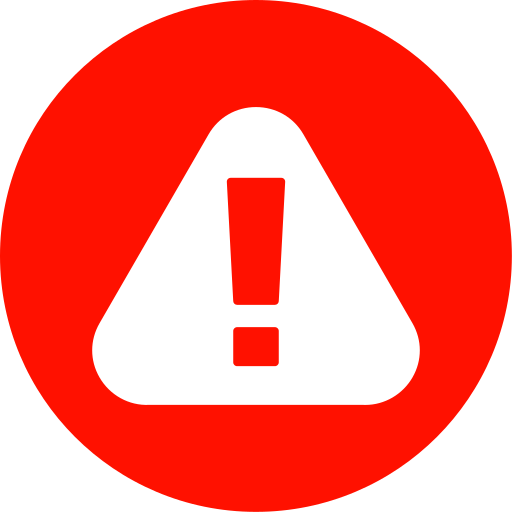}}{}}
\NewDocumentCommand\emojiwarning{}{{\includegraphics[scale=0.02]{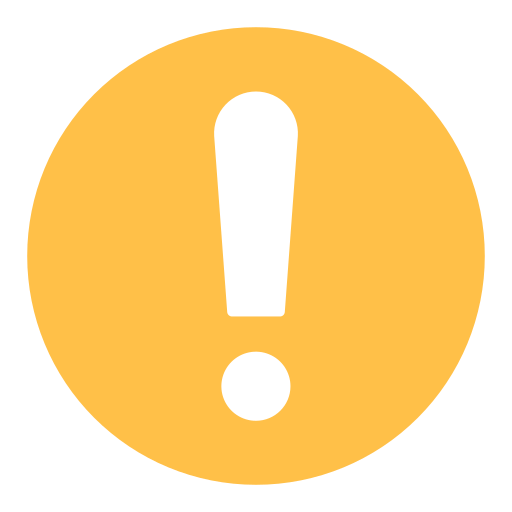}}{}}

\NewDocumentCommand\emojiok{}{{\includegraphics[scale=0.025]{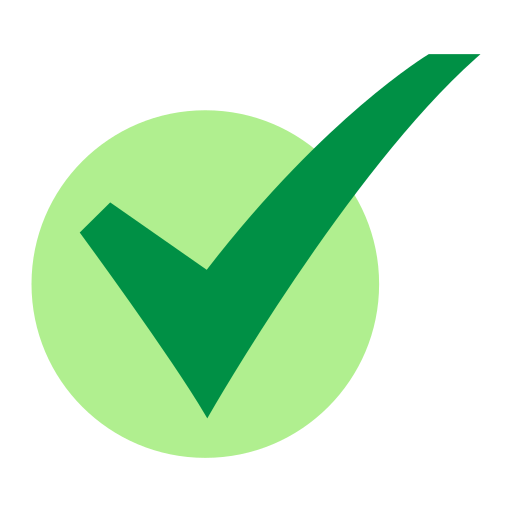}}{}}

\NewDocumentCommand\emojiagree{}{\scalerel*{\includegraphics{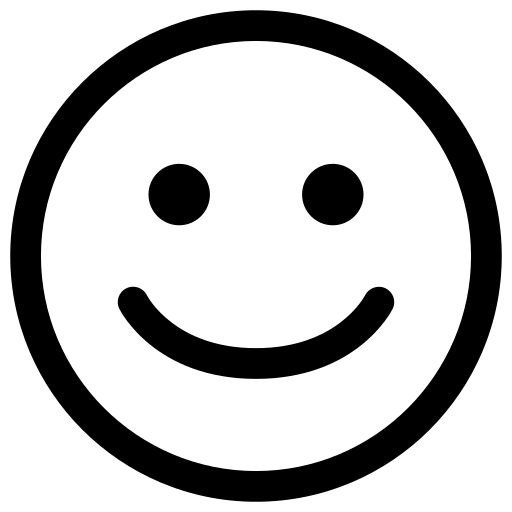}}{X}}
\NewDocumentCommand\emojiagreemost{}{\scalerel*{\includegraphics{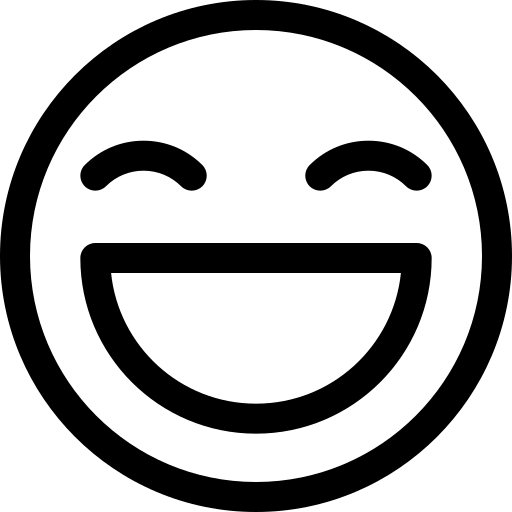}}{X}}
\NewDocumentCommand\emojineutral{}{\scalerel*{\includegraphics{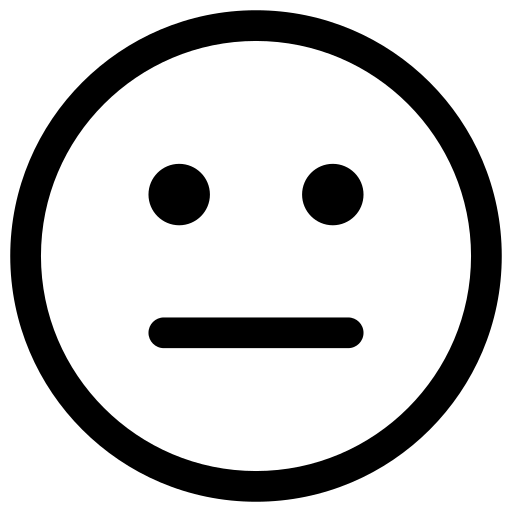
}}{X}}
\NewDocumentCommand\emojidisagree{}{\scalerel*{\includegraphics{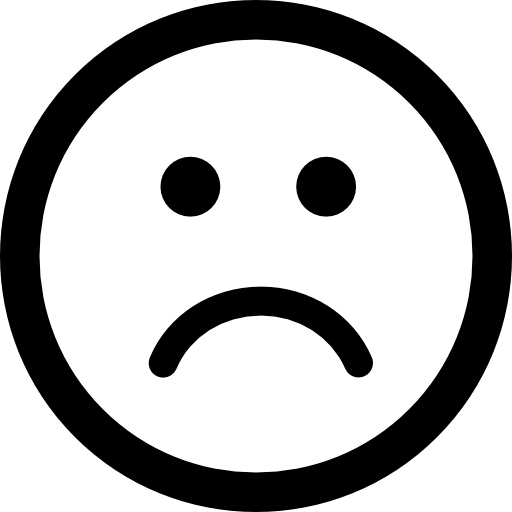
}}{X}}
\NewDocumentCommand\emojidisagreemost{}{\scalerel*{\includegraphics{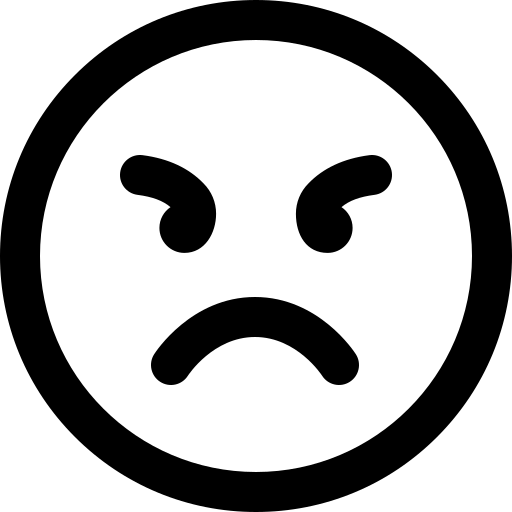
}}{X}}

\usepackage{amssymb}
\usepackage{pifont}%
\newcommand{\cmark}{\ding{51}}%
\newcommand{\xmarkk}{\color{black}\ding{55}}%

\usepackage{bm}

\usepackage{comment}
\usepackage{microtype}
\DeclareMathOperator*{\argmax}{arg\,max}

\usepackage{fixltx2e}

\usepackage{array}
\usepackage{colortbl}

\usepackage{url}
\usepackage{subcaption}

\usepackage{pgfplots}
\usepackage{contour}
\usepackage{xspace}
\usepackage{paralist}
\pgfplotsset{width=10cm,compat=1.9} 

\usepackage{algorithm,algpseudocode}
\usepackage{amsmath}
\usepackage{amsfonts}
\usepackage{amssymb}
\usepackage{color,soul}

\definecolor{darkblue}{rgb}{0,0,.5}
\definecolor{darkgreen}{rgb}{0,.7,0}
\definecolor{lightgray}{rgb}{.8,.8,.8}
\definecolor{aliceblue}{rgb}{0.75, 0.75, 1.0}
\definecolor{darkseagreen}{rgb}{0.46, 0.74, 0.46}
\definecolor{alizarin}{rgb}{0.82, 0.1, 0.26}
\definecolor{airforceblue}{rgb}{0.36, 0.54, 0.66}
\definecolor{red_graph}{rgb}{0.98, 0.8, 0.8}
\definecolor{blue_graph}{rgb}{0.8, 0.98, 0.8}
\definecolor{red}{rgb}{0.8, 0.0, 0.0}
\definecolor{burgundy}{rgb}{0.5, 0.0, 0.13}
\definecolor{britishracinggreen}{rgb}{0.0, 0.26, 0.15}
\definecolor{blue1}{rgb}{0,0,.5}
\definecolor{blue2}{rgb}{0,.7,0}
\definecolor{blue3}{rgb}{.93,.93,.93}
\definecolor{gneon}{rgb}{0.80, 0.9, 0.99}
\definecolor{yneon}{rgb}{0.7, 1.0, 0.06}
\usepackage[inline]{enumitem}
\usepackage{xcolor}

\newcommand{\hname}{contrastive phrasal\xspace}

\newcommand{\cfbox}[2]{%
    \colorlet{currentcolor}{.}%
    {\color{#1}%
    \fbox{\color{currentcolor}#2}}%
}

\newcommand{\ccfbox}[2]{%
    \colorlet{currentcolor}{.}%
    {\color{#1}%
    \fbox{\color{currentcolor}#2}}%
}

\newcommand{\nh}{\textsc{\cfbox{red}{\xmarkk\xspace highlights\hspace{0.1em}}}\xspace}
\newcommand{\yh}{\textsc{\ccfbox{blue}{\cmark\xspace highlights}}\xspace}
\definecolor{celadon}{rgb}{0.67, 0.88, 0.69}
\definecolor{amber}{rgb}{1.0, 0.75, 0.0}

\newcommand{\major}{\emojicancel}
\newcommand{\minor}{\emojiwarning}
\newcommand{\noerror}{\emojiok}

\let\emptyset\varnothing

\newcommand{\dataviza}[1]{\begin{tikzpicture}[scale=2.]
\draw (#1/8 + 0.1,-0.06) node[anchor=south] {\textcolor{celadon}{$\mathbf{#1}$}};
\filldraw[draw=celadon, fill=celadon] (0,0) rectangle (#1/8,0.15); 
\end{tikzpicture}}

\newcommand{\datavizha}[1]{\begin{tikzpicture}[scale=2.]
\draw (#1/8 + 0.1,-0.06) node[anchor=south] {\textcolor{darkgreen}{$\mathbf{#1}$}};
\filldraw[draw=darkgreen, fill=darkgreen] (0,0) rectangle (#1/8,0.15);
\end{tikzpicture}}

\newcommand{\datavizn}[1]{\begin{tikzpicture}[scale=2.]
\draw (#1/8 + 0.1,-0.06) node[anchor=south] {\textcolor{amber}{$\mathbf{#1}$}};
\filldraw[draw=amber, fill=amber] (0,0) rectangle (#1/8,0.15); 
\end{tikzpicture}}

\newcommand{\datavizd}[1]{\begin{tikzpicture}[scale=2.]
\draw (#1/8 + 0.1,-0.06) node[anchor=south] {\textcolor{orange}{$\mathbf{#1}$}};
\filldraw[draw=orange, fill=orange] (0,0) rectangle (#1/8,0.15);
\end{tikzpicture}}

\newcommand{\datavizhd}[1]{\begin{tikzpicture}[scale=2.]
\draw (#1/8 + 0.1,-0.06) node[anchor=south] {\textcolor{alizarin}{$\mathbf{#1}$}};
\filldraw[draw=alizarin, fill=alizarin] (0,0) rectangle (#1/8,0.15);
\end{tikzpicture}}

\contourlength{0.2pt}
\newcommand{\hz}{\vphantom{\parbox[c]{0.1cm}{\rule{0.1cm}{0.4cm}}}}

\usepackage{color, colortbl}
\newcolumntype{o}{>{\columncolor{gray!10}}c}
\newcolumntype{m}{>{\columncolor{gray!30}}c}
\newcolumntype{h}{>{\columncolor{gray!50}}c}

\usepackage{xcolor}

\newcommand{\ignore}[1]{}
\title{
Explaining with Contrastive Phrasal Highlighting: \\
A Case Study in Assisting Humans to Detect Translation Differences}

\author{Eleftheria Briakou$^1$\thanks{~\ Work done while at the University of Maryland.}, \ \   Navita Goyal$^2$,  Marine Carpuat$^2$ \\
  $^1$ Google,
  $^2$ University of Maryland \\
 \texttt{{ebriakou@google.com}},
 \texttt{\{navita, marine\}@cs.umd.edu}
 } 

\begin{document}
\maketitle

\begin{abstract}
Explainable \textsc{nlp} techniques primarily explain by answering \textit{``Which tokens in the input are responsible for this prediction?''}. We argue that for \textsc{nlp} models that make predictions by comparing two input texts, it is more useful to explain by answering \textit{``What \underline{differences} between the two inputs explain this prediction?''}.
We introduce a technique to generate \textit{\hname highlights} that explain the predictions of a  semantic divergence model via phrase-alignment-guided erasure. We show that the resulting highlights match human rationales of cross-lingual semantic differences better than popular post-hoc saliency techniques and that they successfully help people detect fine-grained meaning differences in human translations and critical machine translation errors. 
\end{abstract}

\section{Introduction}

\begin{figure}[t]
    \centering
    \includegraphics[scale=0.27]{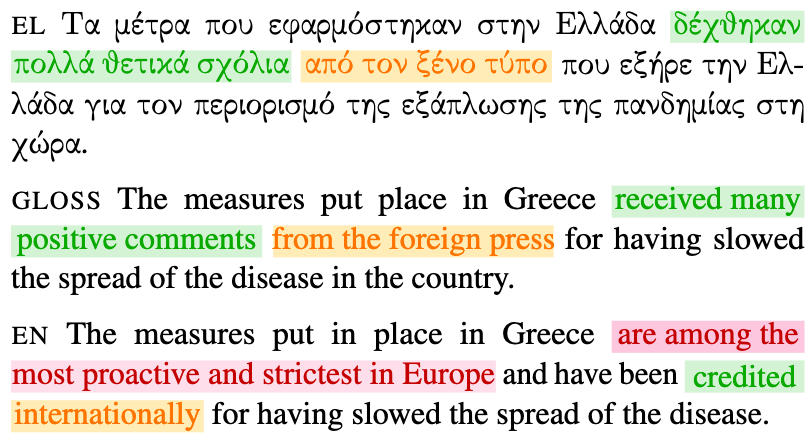}
   \caption{The color-coded \hname highlights explain meaning differences between a Greek sentence (\textsc{el}) and its English translation (\textsc{en}) from Wikipedia. }\label{fig:ex}\vspace{-1em}
\end{figure}

A common strategy to explain the predictions of \textsc{nlp} models is to highlight salient tokens in their inputs~\cite{Li2016UnderstandingNN, lime, NIPS2017_8a20a862}. However, this is suboptimal for the many \textsc{nlp} tasks that require comparing and contrasting two or more pieces of text to predict a class or a similarity score, such as natural language inference~\cite{bowman-etal-2015-large}, 
semantic textual similarity~\cite{agirre-etal-2012-semeval}, or evaluation and quality estimation of text generation~\cite{bojar-etal-2017-results, ma-etal-2018-results, ma-etal-2019-results, mathur-etal-2020-results, freitag-etal-2021-results, freitag-etal-2022-results}. 
At the same time, a long line of research in social sciences shows that human explanations are contrastive~\cite{MILLER20191}, i.e., humans do not explain why an event happened in a vacuum but instead compare what happened to a contrast case. For instance, to explain how a Greek sentence and a translation in English differ, it is more intuitive to show how they differ (as presented in Figure~\ref{fig:ex}) than to highlight all salient Greek and English tokens without specifying how they relate to each other. 

In this work, we introduce a post-hoc explainability technique to answer  \textit{``What differences between two inputs explain a prediction''?} for \textsc{nlp} regression models that produce a score encoding a relationship between two text inputs. We apply it to the task of explaining meaning divergences between a text in one language and its translation. We design a phrase-alignment-guided erasure strategy to explain the predictions of a semantic divergence model and highlight minimal phrase pairs in the two inputs that result in meaning differences.

We extensively evaluate our approach with both proxy and human-centered evaluations. 
Comparing against human explanations for meaning differences in English-French human translations, our approach matches human rationales better than popular post-hoc methods that highlight salient tokens independently~\cite{lime, NIPS2017_8a20a862}. 
Acknowledging the limitations of such proxy evaluations \citep{Buccinca2020ProxyTA}, we conduct two \textsc{irb}-approved human studies to assess the usefulness of our \hname highlights when crowd-sourcing challenging bilingual annotation tasks. We find that our approach helps annotators reliably discover fine-grained meaning differences in English-French and English-Spanish human translations and detect critical errors in Portuguese-to-English machine translation.  We make our code and data publicly available: \url{https://github.com/Elbria/Ex-SemDiv}.
\section{Background}

We focus on \textit{local} explanations which explain why a specific prediction was made, often to help humans rely on predictions appropriately in machine-in-the-loop use cases~\cite{doshi2017towards}.

\paragraph{Highlight Explanations}

Local explanations often consist of highlighting input features deemed important to a prediction.
In glass-box settings, prior works seek to quantify feature saliency with model-internal signals such as gradients \cite{Arras2016WhatIR, Smilkov2017SmoothGradRN}, contextual decompositions \cite{murdoch2018beyond}, or attention mechanisms \cite{Choi2016RETAINAI}. However, these highlights' ability to serve as explanations has been questioned~\cite{Jain2019AttentionIN, wang-etal-2020-gradient, moradi-etal-2021-measuring}, leading to work revisiting technical solutions and evaluations under which 
attention \textit{could} be seen as an explanation~\cite{wiegreffe-pinter-2019-attention, moradi-etal-2019-interrogating, tutek-snajder-2020-staying}. 
Another family of approaches relies on erasure strategies to explain the predictions of black box models. \citet{Li2016UnderstandingNN} quantify a token's importance as the difference in model confidence once it is erased from the input, with several following works exploring different token erasure schemes~\cite{feng-etal-2018-pathologies, kim-etal-2020-interpretation}.
\textsc{lime}~\cite{lime} approximates the model's local decision boundary with linear models by erasing multiple tokens from a model's input, while \textsc{shap}~\cite{NIPS2017_8a20a862} computes Shapley values by estimating the marginal contribution of each token across all possible erasures.
We build on this line of work by designing an erasure strategy that identifies salient \textit{\hname pairs} rather than independent salient tokens.

\paragraph{Contrastive Explanations}
Motivated by social science research suggesting that humans explain with respect to an (implicit) contrast case~\cite{MILLER20191}, recent work has sought to \textit{generate} the latter by producing counterfactual explanations: minimally edited versions of the input that change the model prediction. 
Performing such edits usually requires training dedicated editors~\cite{ross-etal-2021-explaining}, prompting language models~\cite{paranjape-etal-2021-prompting}, and accessing knowledge bases~\cite{chen-etal-2021-kace}, among others~\cite{li-etal-2020-linguistically, chemmengath-etal-2022-cat}. Closer to our work, another family of approaches extracts \textit{highlights} contrastively: ~\citet{jacovi-goldberg-2021-aligning} extend erasure-based approaches to identify which input features lead to the actual prediction vs. a contrastive prediction, while \citet{yin-neubig-2022-interpreting} extend saliency methods to explain the generation of contrast cases for language generation tasks. 
Our work explains the predictions of models that compare two inputs, where identifying salient phrase pairs that differ across them provides a natural mechanism for contrastive explanations.

\paragraph{Evaluation} 
Most current work in \textsc{nlp} adopts \textit{proxy} evaluations that compare automatic explanations with human explanations of the gold label, with numerous datasets encoding explanations using highlights or free-form text~\cite{698d51a1}.
Others rely on \textit{simplified} human tasks, such as simulatability, evaluating whether explanations help people predict a model's prediction more accurately~\cite{hase-bansal-2020-evaluating, nguyen-2018-comparing}. 
Despite the attractiveness of the above evaluations, evidence is growing that they do not reliably indicate how useful explanations are in practice~\cite{Buccinca2020ProxyTA}. \citet{boyd-graber-etal-2022-human} call for application-grounded evaluations to directly assess whether explanations help people complete an actual task, as suggested by work in \textsc{hci}~\cite{10.1145/3411764.3445088, Liao2020QuestioningTA}. 
We heed that call by complementing a proxy evaluation with two application-grounded user studies. 

\paragraph{Explaining Translation Differences} 
Detecting translation differences is a core task in multilingual \textsc{nlp}, e.g., to predict machine translation errors \cite{rei-etal-2020-comet}, to study how humans translate \cite{zhai-etal-2020-detecting}, or to understand how multilingual resources such as Wikipedia differ across languages~\cite{10.3389/fphy.2018.00054, JIANG2017248}. 
Automatic approaches often score the degree of (dis)similarity between an input and its translation to quantify machine translation quality~\cite{Zhang2019BERTScoreET, sellam-etal-2020-bleurt, rei-etal-2020-comet} or to detect meaning divergences in human translations~~\cite{vyas-etal-2018-identifying, briakou-carpuat-2020-detecting, wein-schneider-2021-classifying}. However, sentence-level scores can be hard to interpret and do not pinpoint how translations differ. 
This has been addressed by tagging text with human translation processes~\cite{zhai-etal-2018-construction}, with dedicated, supervised models or with word-level quality estimation (\textsc{qe}) tasks~\cite{specia-etal-2018-findings, specia-etal-2020-findings-wmt, specia-etal-2021-findings}, often addressed by applying post-hoc explanation methods to sentence-level predictors \cite{treviso-etal-2021-ist, rei-etal-2022-cometkiwi}. We augment this work by contributing unsupervised contrastive explanations of translation differences and testing whether they help annotate them more reliably.

\begin{figure*}[h]
    \centering
    \includegraphics[width=0.99\linewidth]{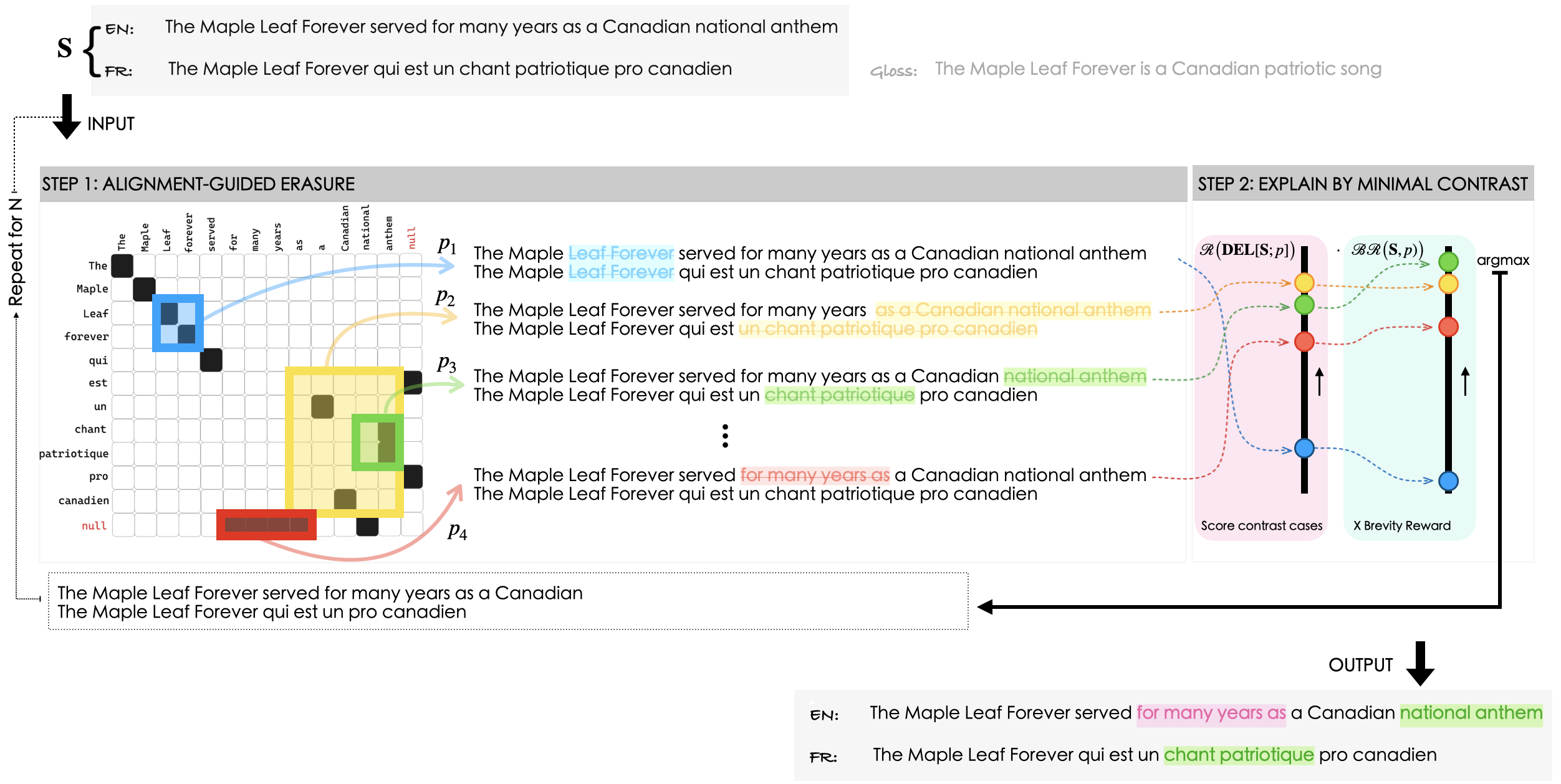}
    \caption{Our approach takes as input a sentence pair ($\mathbf{S}$) and extracts a set of perturbed inputs by erasing phrasal pairs guided by word alignments (Step 1). Then it explains the prediction of a regressor $\mathcal{R}(\mathbf{S})$ by highlighting the  phrasal pair $p$ that,  once deleted, maximizes the model's prediction $\mathcal{R}(\mathrm{\textsc{\textbf{del}}}[\mathbf{S};p])$ multiplied by a brevity reward $\mathcal{BR}(\mathbf{S}, p)$  that encourages the extraction of short phrasal pairs (Step 2).}
    \label{fig:approach}
\end{figure*}
\section{Explaining Divergences with Contrastive Phrasal Highlights}

We introduce a highlighting approach to explain the predictions of a model that compares and contrasts two text inputs.
We hypothesize that phrase pairs provide more appropriate \textit{cognitive chunks}~\cite{doshi2017towards} than tokens as basic units of explanations. 
For instance,  instead of highlighting tokens that differ in the English-French pair of Figure~\ref{fig:approach}, it is more natural to explain that the English text refers to the Maple Leaf Forever as a \textit{national anthem} as opposed to a \textit{chant patriotique} (patriotic song) in French and that the English phrase \textit{``for many years''} has no equivalent in French.

Building on token-level erasure-based explanation methods~\cite{Li2016UnderstandingNN}, we design a phrase-alignment-guided erasure strategy to explain the predictions of a model \cite{briakou-carpuat-2020-detecting} that ranks bilingual sentence pairs based on the granularity of their semantic similarity 
(assuming $\mathbf{S}$ is an equivalent pair and $\hat{\mathbf{S}}$ a pair containing semantic divergences, the model $\mathcal{R}(\cdot)$ ranks them such that $\mathcal{R}(\mathbf{S}) > \mathcal{R}(\hat{\mathbf{S}})$).
As shown in Figure~\ref{fig:approach}, given two input texts, we first extract a set of candidate counterfactual inputs by masking a subset of phrase pairs (\S\ref{sec:alignment_guided}). Then, we explain the model's prediction by selecting the phrase pair whose erasure maximally increases the similarity score between the two inputs (\S\ref{sec:minimal_contrast}).

\subsection{Alignment-Guided Phrasal Erasure}\label{sec:alignment_guided}

We propose an erasure approach that takes into account the input's structure. Given a sentence pair $\mathbf{S}$, we start by extracting a set of candidate counterfactual instances by deleting a single phrase pair from $\mathbf{S}$. Given that erasing all possible phrase pairs for each sentence is computationally prohibitive, we restrict our search to deleting phrases that belong to an aligned phrase table, $\mathcal{P}$. 

Our phrase pair candidates for erasure are derived from classical statistical machine translation techniques developed for extracting phrase translation dictionaries from parallel corpora~~\cite{och-etal-1999-improved, koehn-etal-2007-moses}. 
Given two texts, we derive word alignments based on multilingual word embeddings~\cite{jalili-sabet-etal-2020-simalign} and then extract all phrase pairs consistent with the word alignments.
A phrase pair $(p_1,p_2)$ is consistent with the word alignment if $p_1$ and  $p_2$ are each contiguous sequences of words in each language, where all alignment links that start within $p_1$ map to tokens within $p_2$ and vice versa.
Unaligned words (i.e., words aligned to a \texttt{\textcolor{red}{null}} token) can be included in either the source ($p_1$) or the target ($p_2$) phrases. As a result, the extracted phrase pairs comprise not only equivalent 
but also include related but divergent phrase pairs (\textit{``national anthem''} and \textit{``chant patriotique''}) and unaligned phrase (\textit{``served for many years''}), as seen in Figure~\ref{fig:approach}.

\subsection{Explaining by Minimal Contrast}\label{sec:minimal_contrast}

Given the sentence pair $\mathbf{S}$ and its aligned phrase table $\mathcal{P}$, we first extract a set of contrast cases $\mathcal{\tilde{P}}$ consisting of all the phrasal pairs that, once erased,
make the two inputs more equivalent, as measured by an increase in score larger than a margin $\epsilon$:

\[
    \mathcal{\tilde{P}} = \Big\{ {p \in \mathcal{P}}, \ \ \text{s.t.}  \ \ \mathcal{R}\Big(
    \mathrm{\textbf{\textsc{del}}}\big[ \mathbf{S};  p\big]
    \Big) > \mathcal{R}\Big( \mathbf{S} \Big) + \epsilon \Big\}
\]

\noindent

Since explanations should be minimal~\cite{Hilton, lipton_1990}, covering only the most relevant causes to reduce cognitive load, we select as \hname highlights the phrase pairs that correspond to minimal edits with a large impact on predictions, as follows:

\[
    \argmax \Bigg\{\mathcal{R}\Big(
    \mathrm{\textbf{\textsc{del}}}\big[ \mathbf{S};  p\big]
    \Big)
    \cdot
    \mathcal{BR}\Big(\mathbf{S}, p\Big)\Bigg\}
\]

\[
    \mathcal{BR}(\mathbf{S}, p)= 
\begin{cases}
    e^{-\frac{|p|}{|\mathbf{S}|}},& \text{if } 
    \mathcal{R}\Big(
    \mathrm{\textbf{\textsc{del}}}\big[ \mathbf{S};  p\big] \Big) \geq 0  \\
    e^{+\frac{|p|}{|\mathbf{S}|}},              & \text{otherwise}
\end{cases}
\]

where the first term $\mathcal{R}(
    \mathrm{\textbf{\textsc{del}}}\big[ \mathbf{S};  p\big]
)$ encourages the extraction of a \hname highlight $p$ corresponding to a high score under the model (i.e., deleting this phrase pair yields a contrast case), while the second term, $\mathcal{BR}\big(\mathbf{S}, p\big)$ corresponds to a \textit{brevity reward} that encourages extraction of shorter highlights. $|\mathbf{S}|$ is computed by adding the length of each of the two sentences, and $\mathrm{\textbf{\textsc{del}}}[\cdot]$ is a function that erases a phrase pair $p$ from $\mathbf{S}$.
\\  

\noindent
The above approach identifies a single phrase pair that explains the divergence prediction for the original sentence $\mathbf{S}$. To extract multiple explanations, we repeat this process by iteratively erasing 
the extracted \hname highlight from the current input sentence
$\mathbf{S'}= \mathrm{\textbf{\textsc{del}}}\big[ \mathbf{S};  p\big]$, 
and repeat the steps
in \S\ref{sec:alignment_guided} and \S\ref{sec:minimal_contrast}.
This iterative process ends when, for a given input, 
none of the extracted counterfactual instances yield a more equivalent pair, i.e., $\tilde{\mathcal{P}} = \emptyset$, or we reach an equivalent input under the divergent ranker, i.e., 
$\mathcal{R}(\mathbf{S'})>0$.
\section{Proxy Evaluation}\label{sec:proxy}

In this section, we describe our proxy evaluation based on human-provided highlights. We acknowledge that proxy evaluations encounter validity issues~\cite{boyd-graber-etal-2022-human} and use them primarily to guide system development and validate against standard highlighting methods.  

\subsection{Experimental Setup}\label{sec:ex_setup_proxy}

\paragraph{Explanandum} We seek to explain the prediction of a divergence ranking model that is trained as recommended in prior work (see Appendix~\ref{sec:appendix_explanandum_details}).
 
\paragraph{Explainers} We contrast our \hname highlights against
a \textsc{random} baseline that highlights tokens in each sentence at random with $0.5$ probability, and 
two standard post-hoc explanation methods, \textsc{lime} and \textsc{shap}, that seek to explain the predictions of the explanandum.

\paragraph{Reference Highlights} We evaluate our approach using the 
\textsc{REFreSD} dataset,\footnote{\url{https://github.com/Elbria/xling-SemDiv/tree/master/REFreSD}} which is manually annotated with divergences of fine-grained and coarse-grained granularity, along with rationales that justify the sentence label. We compare our \hname highlights with the human rationales on a subset of $418$ challenging instances annotated as having  ``Some Meaning Differences'' at the sentence level,  where we expect the more subtle differences in meaning to be found.

\begin{table*}[!t]
    \centering
    \scalebox{0.9}{
    \begin{tabular}{l@{\hskip 0.5in}r@{\hskip 0.1in}r@{\hskip 0.1in}r@{\hskip 0.1in}r@{\hskip 0.5in}r@{\hskip 0.1in}r@{\hskip 0.1in}r@{\hskip 0.1in}r}
    \\\toprule[2pt]
   & \multicolumn{4}{c}{\textbf{\textsc{english}}} & \multicolumn{4}{c}{\textbf{\textsc{french}}}\\
       & \textbf{\textsc{pr.}} & \textbf{\textsc{re.}} & \textbf{\textsc{f}1}   & \textbf{\textsc{del.}}  & \textbf{\textsc{pr.}} & \textbf{\textsc{re.}}  &  \textbf{\textsc{f}1} & \textbf{\textsc{del.}}  \\

    \addlinespace[0.2cm]
    \rowcolor{gray!10}

    \textsc{oracle} & $100$ & $100$ & $100$ &$39\%$ &  $100$ & $100$ & $100$ & $43\%$ \\
    
    \textsc{random}                  & $39$ & $48$ & $43$ & $50\%$ & $42$ & $49$ & $45$ & $50\%$  \\
    
    \textsc{lime}   & $45$ & $37$ & $41$ & $30\%$ & $44$	& $37$ & $40$ &  $34\%$  \\ 
    \textsc{shap}  & $53$ & $34$ & $41$ & $25\%$ & $50$ & $32$ & $39$ & $26\%$  \\
    \cmidrule{2-8}
    
    \textsc{ours} ($-$ BR)                    & $52$ & $76$ & $62$ & $54\%$ & $56$ & $74$ & $64$ & $55\%$ \\
    \textsc{ours} & $58$ & $61$ & $62$ & $37\%$ & $60$ & $55$  & $64$ & $37\%$  \\ 

    \toprule[2pt]
    \end{tabular}}
    \caption{Proxy evaluation results with respect to human rationales on \textsc{REFreSD}. 
    }
    \label{tab:proxytable}\vspace{1em}
\end{table*}

{
\renewcommand{\arraystretch}{1.3}
\begin{table*}[!ht]
    \centering
    \scalebox{0.54}{
    \begin{tabular}{ll}
    \toprule[3pt]
    \addlinespace[0.2cm]
    
    \multirow{3}{*}{\textbf{\textsc{human}}}
     & The older generation \colorbox{yellow!50}{\hz turbines generate kilowatts , and the modern turbines installed generate up to} 3 megawatts , \colorbox{yellow!50}{\hz depending on the specific turbine and manufacturer ..}\\
     & Les turbines d' ancienne génération génèrent des kilowatts , alors que les \colorbox{yellow!50}{\hz éoliennes} modernes ont \colorbox{yellow!50}{\hz une puissance pouvant aller jusqu' à} 3 mégawatts .\\
     & \it (gloss) The turbines of the old generation generate kilowatts, while the modern wind turbines generate up to 3 megawatts of power.\\

    \addlinespace[0.2cm]
    \cline{2-2}
    \addlinespace[0.2cm]
    
    \multirow{2}{*}{\textbf{\textsc{lime}}}
    & \colorbox{yellow!50}{\hz The} older \colorbox{yellow!50}{\hz generation  turbines} generate kilowatts , and the modern turbines \colorbox{yellow!50}{\hz installed generate} up to 3 megawatts , \colorbox{yellow!50}{\hz depending on} the \colorbox{yellow!50}{\hz specific} turbine \colorbox{yellow!50}{\hz and manufacturer ..}\\
    & \colorbox{yellow!50}{\hz Les turbines d' ancienne génération génèrent des} kilowatts  \colorbox{yellow!50}{\hz,} alors que les éoliennes modernes ont une puissance pouvant  \colorbox{yellow!50}{\hz aller} jusqu' à 3 mégawatts .\\

    \addlinespace[0.2cm]
    \cline{2-2}
    \addlinespace[0.2cm]
    
    \multirow{2}{*}{\textbf{\textsc{shap}}}
    & The older generation turbines generate kilowatts , \colorbox{yellow!50}{\hz and} the modern turbines installed generate up to 3 megawatts \colorbox{yellow!50}{\hz ,} depending \colorbox{yellow!50}{\hz on} the specific \colorbox{yellow!50}{\hz turbine and manufacturer} ..\\
    & Les turbines d' ancienne génération génèrent des kilowatts , alors que les éoliennes modernes ont une puissance pouvant aller jusqu' à 3 mégawatts .\\

    \addlinespace[0.2cm]
    \cline{2-2}
    \addlinespace[0.2cm]

    \textbf{\textsc{ours}} 
     & The older generation turbines generate kilowatts , and the modern turbines \colorbox{gneon}{\hz \textcolor{airforceblue}{\textbf{installed generate}}} up to 3 megawatts, \colorbox{yneon}{\hz \textcolor{darkgreen}{\textbf{depending on the specific turbine and manufacturer ..}}}\\
     & Les turbines d' ancienne génération génèrent des kilowatts , alors que les éoliennes modernes \colorbox{gneon}{\hz \textcolor{airforceblue}{\textbf{ont une puissance}}} pouvant aller jusqu' à 3 mégawatts .\\

    \addlinespace[0.2cm]
    \toprule[3pt]

    \end{tabular}} 
    \caption{Examples of divergence explanations (\textsc{\textbf{human}} corresponds to \textsc{REFreSD} rationales).} 
  \label{tab:refresd_examples}
\end{table*}
}

\paragraph{Evaluation Metrics} Following prior proxy evaluations of explanations~\cite{ross-etal-2021-explaining, deyoung-etal-2020-eraser},
we compute:
\begin{inparaenum}
    \item \textit{Agreement with Human Rationales}: Precision, Recall, and F-1 scores 
    computed against human rationales;
    \item \textit{Minimality}: the length of the \hname highlights, measured as the number of highlighted tokens.
\end{inparaenum}

\subsection{Results}

As seen in Table~\ref{tab:proxytable}, explaining divergence predictions by extracting \hname highlights significantly outperforms both \textsc{lime} and \textsc{shap}.
Those standard highlighting baselines even underperform the \textsc{random} baseline.\footnote{Since \textsc{lime} and \textsc{shap} are feature attribution methods that assign a continuous score to each word, we checked that changing the threshold from the default $t=0$ does not improve the results, as both methods suffer from a precision-recall tradeoff.} 
A closer look at the outputs (Table~\ref{tab:refresd_examples}) indicates that both \textsc{lime} and \textsc{shap} suffer from two major issues: sparsity and accuracy. We attribute those issues to the fact that both approaches operate in token space, ignoring the interdependent relationships between tokens across the two languages. By explicitly modeling such relationships, our approach produces highlights that match the ones in \textsc{REFreSD} better.

We measure the impact of the brevity reward by ablation (i.e., \textsc{ours} ($-$ \textsc{br})). This achieves the highest recall at the expense of precision: dropping the brevity reward produces fewer and longer highlights per instance. It highlights more than $50\%$ of each sentence pair on average, while humans only highlight about $40\%$. The brevity reward thus helps match reference highlights better, supporting the benefits of producing minimal explanations.\\

\noindent
In sum, these results are promising indicators that \hname highlights could be helpful when detecting semantic divergences. However, as highlighted by recent work on explanation evaluation, proxy evaluations can be misleading. 
Given that prior work evaluates highlight-based explanations of (machine) translation errors based \textit{solely on automatic evaluations}, we chose to initiate a human-centered discussion that evaluates the usefulness of such highlights directly with an in-depth exploration of multiple small-scale \textit{application-grounded} evaluations as detailed below.
\section{Application-Grounded Evaluation I: Annotation of Semantic Divergences}\label{sec:annotation}

As shown by prior work, annotating fine-grained divergences is a challenging task and requires dedicated annotation protocols based on human rationales~\cite{briakou-carpuat-2020-detecting} or abstract meaning representation frameworks~\cite{wein-2022-spanish} to achieve moderate agreement. 
As a result, such annotations are usually hard to collect with crowd workers since they require explicit annotator training. 
In this section, we ask whether \hname highlights help crowd-workers annotate divergences more reliably, which could ease the need for complex and expensive annotation protocols. Concretely, we test the following hypothesis: \textit{Contrastive phrasal highlights improve the annotation of fine-grained semantic divergences in terms of accuracy and agreement}.
 
\begin{figure*}[!t]
\begin{subfigure}{.5\textwidth}
  \centering
  \includegraphics[width=.9\linewidth]{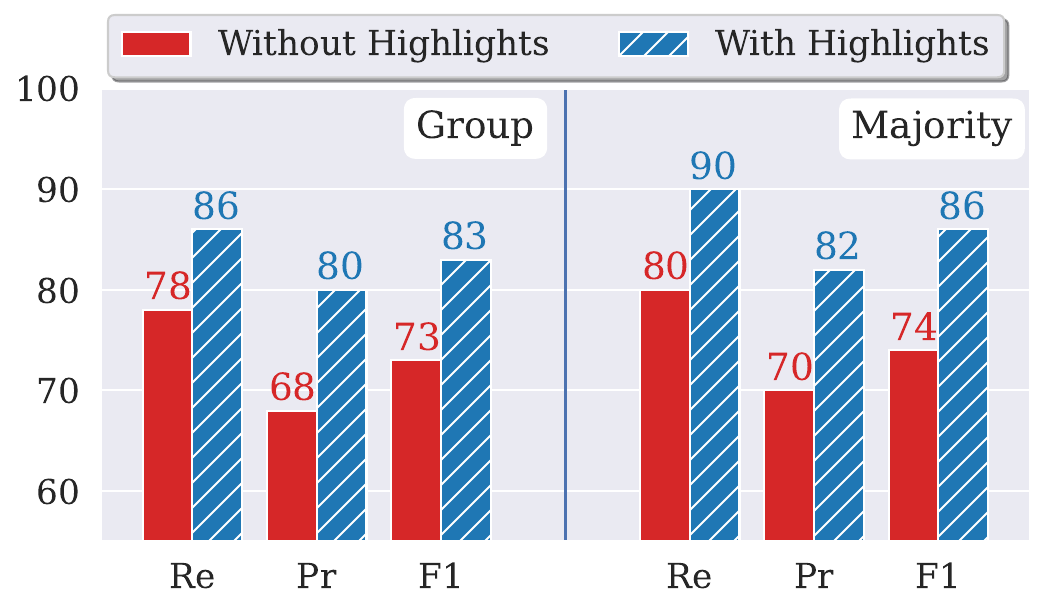}
  \caption{English-French}
  \label{fig:sfig1}
\end{subfigure}
\begin{subfigure}{.5\textwidth}
  \centering
  \includegraphics[width=.9\linewidth]{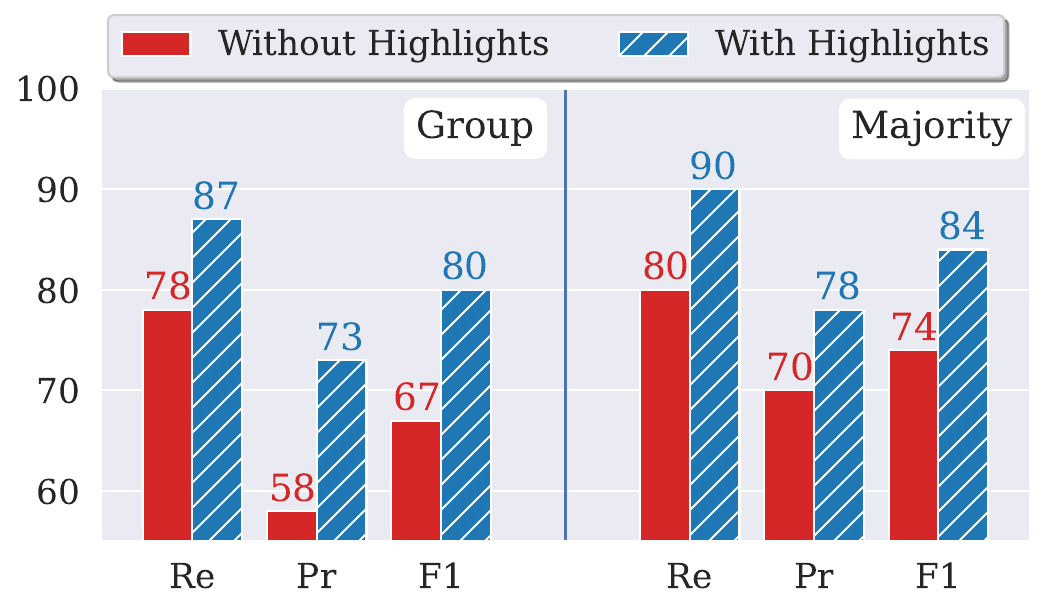}
  \caption{English-Spanish}
  \label{fig:sfig2}
\end{subfigure}
\caption{Annotation-grounded evaluation comparing with- vs. without-highlights annotation conditions. Contrastive phrasal highlights significantly ($p=0.1$) improve the annotation of fine-grained semantic divergences.}
\label{fig:reliability_application_grounded}
\end{figure*}

\subsection{Experimental Setup}\label{sec:ex_setup_user_1}

\paragraph{Explanandum} We seek to explain the predictions of Divergent m\textsc{bert}. We train separate models for English-French and English-Spanish following the approach detailed in Appendix~\ref{sec:appendix_explanandum_details}.

\paragraph{Study Data} 

To provide controlled yet realistic test samples, we construct a dataset of fine-grained divergences
that mimic translation processes used by human translators~\cite{Zhai2019TowardsRP}. We start with English-French and English-Spanish human translations from the multi-parallel \textsc{flores} benchmark~\cite{goyal-etal-2022-flores}.
We randomly select $50$ \textit{seed} translations among those identified as semantically equivalent pairs by Divergent m\textsc{bert}. Then, we introduce fine-grained meaning differences by editing the English references of $10$ of those samples.  We perform edits motivated by translation processes, such as modulation, explicitation, and reduction~\cite{Zhai2019TowardsRP}, by interaction with Chat\textsc{gpt}.\footnote{Details of the Chat\textsc{gpt} prompts are in Appendix~\ref{sec:chatGPT_details}.}
For instance, we introduce a generalization as follows:\\

\begin{tabular}{p{0.9\columnwidth}}
\textcolor{gray}{Despite leaving the show in $1993$ he kept}  \textit{[the title of executive producer]} $\rightarrow$ \textit{[a senior role]}  \textcolor{gray}{and continued to receive tens of millions of dollars every season in royalties.}\\
\end{tabular}\\

To create difficult examples of equivalent translations, we also paraphrase the English references of 10 seed examples with Chat\textsc{gpt} to introduce syntactic changes. We validated all examples, ensuring that edits actually match the intended divergences in the first case (Figures~\ref{fig:interface_examples_added_1} and \ref{fig:interface_examples_changed_1}) and that syntactic changes do not alter meaning in the second (Figure~\ref{fig:interface_examples_paraphrased_1}).
The final dataset consists of $35$ semantically equivalent translations (split into $25$ original translations and $10$ syntactically divergent paraphrases) 
and $15$ fine-grained divergences resembling translation processes. 


\subsection{Study Design}\label{sec:study_design_1}

We ran a controlled evaluation across the two language pairs (English-French and English-Spanish). We describe the task and 
study design below.

\paragraph{Task Description} 

An annotation instance is a sentence pair in English and, e.g., French. Bilingual speakers are asked to read each sentence pair closely, determine whether the \textit{``sentence pair contains any difference in meaning''}, and finally characterize the difference as ``Added'' vs. ``Changed''.
``Added'' 
refers to added information, while ``Changed'' refers to information that is present in both languages but does not precisely match---mirrorring prior annotation divergence protocols~\cite{briakou-carpuat-2020-detecting}. We include a screenshot of the task description presented to annotators in Figures \ref{fig:instructions_1} and \ref{fig:interface_examples_changed_1} of Appendix \ref{sec:appendix_interfaces}.

\paragraph{Conditions}
We study two conditions: one in which participants are shown highlights (\yh) and a second in which they are not  (\nh).
The only information available to participants about the highlights is that they \textit{``are AI-generated and indicative of meaning differences''}. 
We emphasize that highlights are not presented as explanations of divergence predictions, as we want annotators to choose how and whether they use them in their assessments based on their intuitions.

\paragraph{Procedures} 

We conduct a between-subjects study where participants are randomly assigned to a condition.
Participants are first presented with a tutorial that explains the task and relevant terminology. 
Each participant is presented with $25$ instances from one of the two studied conditions. 
Each batch of annotated instances contained ~$30\%$ of semantically divergent edits, ~$20\%$ of syntactically divergent edits, and ~$50\%$ original semantically equivalent pairs. Instances within each batch are randomized. We include two attention checks where participants are asked to indicate their answers to the previous question.
After completing the task, participants were asked to complete a brief survey assessing their perception of the generated highlights (if assigned to this condition) and finally were asked to provide their free-form feedback on the study along with demographic information, such as gender and age.
The average time of the study was ~$20$ minutes.
In sum, we collect $3$ annotations per instance and annotate a total of $100$ instances ($50$ per condition) for each of the two language pairs studied.

\paragraph{Participants} We recruit $12$ participants per language pair using  Prolific.\footnote{\url{https://www.prolific.co/}}
Each participant is restricted to taking the study only once. None of them failed both attention checks; hence we did not exclude any of the collected annotations from our analysis.
All participants identified as bilingual speakers in the languages involved in each study. Participants are compensated at $15$ \textsc{usd} per hour.

\subsection{Measures}\label{sec:annotation_measures}
Our main evaluation measures are Precision, Recall, and F1 computed by assuming that semantically edited instances correspond to divergences, while the rest are treated as semantically equivalent pairs. We report those accuracy statistics both at the group level and also by majority voting annotations. Furthermore, we summarize the responses provided as free-form feedback and report subjective measures that reflect participants' perceived understanding of the explanations, i.e., \textit{``the highlights are useful in detecting meaning differences''}, if provided.
The latter measures are collected on a 5-point Likert scale. 
We report statistical significance based on bootstrap resampling. We draw samples randomly from the collected annotations with replacement. The sample size is the same as the one of the collected annotations, while the number of resamples is set to $1{,}000$. For the application-grounded evaluation I, which contains a total of $150$ annotations per language, the significance level is $p=0.1$.



\subsection{Study Results}\label{sec:results_user_1}

\paragraph{Reliability of Annotations} As shown in Figure~\ref{fig:reliability_application_grounded}, the bilingual \textit{group} that annotated fine-grained meaning differences in the presence of contrastive explanations---\yh, is significantly ($p=0.1$) more accurate across Precision, Recall, and F1 scores and both language pairs, compared to \nh. Furthermore, those improvements carry over when aggregating annotation results by \textit{majority} voting annotations across instances. As a result, this leads us to accept the hypothesis that \hname highlights improve the annotation of fine-grained divergences by bilingual speakers. Finally, detecting divergences with highlights additionally improves the reliability of annotations are measured by Cohen's Kappa inter-annotator agreement statistics: 
\begin{table}[!ht]
    \scalebox{.85}{
    \centering
    \begin{tabular}{lll}
    & \textsc{en-fr} & \textsc{en-es} \\ 
    \nh & $0.51$ (moderate) &  $0.33$ (fair)  \\
    \yh & $0.66$ (substantial) & $0.52$ (moderate)\\
    \end{tabular}}
\end{table}

\paragraph{Subjective Measures \& User Feedback} Overall, bilingual speakers presented with \hname highlights agreed they were useful in helping them spot fine-grained divergences---average self-reported usefulness of $3.8$ for \textsc{en-fr} and $4.2$ for \textsc{en-es}.
Finally, although \hname highlights were useful, annotators also note that they cannot entirely rely on them. For instance, some of the participant's feedback is \textit{``The highlights are useful but not $100\%$ reliable, that is because I found other added/changed words that AI did not highlight'' } and \textit{``In most cases, the words highlighted by the AI have helped to detect possible differences''}.
\section{Application-Grounded Evaluation II: 
Critical Error Detection}

In this section, we examine the potential of using \hname highlights to assist bilingual humans in detecting critical errors in machine translation outputs.
Recent work on quality estimation for machine translation proposes an error-based evaluation framework where bilingual \textit{professional} annotators are asked to: highlight translation errors; rate their severity as major or minor; and finally, indicate their type (e.g., mistranslation vs. terminology, etc.)~\cite{freitag-etal-2021-experts}. 
Drawing on those intuitions, we aim to study whether a simplified  evaluation framework that uses \hname highlights
yields reliable annotations of critical errors with bilingual \textit{crowd workers}.
This is a hard task as these errors are rare in high-quality systems and might be missed when parsing translations quickly.
In what follows, we test the hypothesis: \textit{Contrastive phrasal highlights help bilingual crowd-workers detect critical (accuracy) errors in machine translation outputs}.

\subsection{Experimental Setup}\label{sec:ex_setup_2}

\paragraph{Explanandum} We explain the predictions of Divergent m\textsc{bert} trained for English-Portuguese following the process detail in Appendix~\ref{sec:appendix_explanandum_details}. 
We emphasize that the explanandum is sensitive to detecting any meaning differences and not only critical errors, and therefore, we expect the \hname highlights to surface both minor and major accuracy errors.

\paragraph{Study Data}
For the purposes of the study, we use the synthetic Portuguese-English dataset from the Critical Error Detection task of \textsc{wmt}~\cite{freitag-etal-2021-results}. The dataset consists of pairs sampled from the \textsc{opus} parallel corpus~\cite{tiedemann-2012-parallel} treated as ``not containing a critical error''. Then, the \textsc{wmt} organizers artificially corrupted $10\%$ of those pairs to introduce critical errors that reflect hallucinated content, untranslated content, or mistranslated texts. We will refer to those errors as \textsc{wmt} for the rest of the paper.
Additionally, to make the task hard for humans, we a) filter out \textsc{wmt} critical errors that concern deviation in numbers, time, units, or dates and b) artificially introduce \textit{negation} errors: we edit the English text locally by changing one or two words in such a way that the meaning of the sentence is entirely flipped (e.g., the word \textit{benefits} in an English reference is replaced by the word \textit{harms}). In total, the study dataset consists of $30$ translations that do not contain a critical error (also detected as equivalent under the explanandum), $10$ translations reflecting \textsc{wmt} errors, and $10$ translations reflecting local \textit{negation} errors.

\subsection{Study Design}\label{sec:study_design_2}
We test our hypothesis with a user study of critical error detection in English translations of Portuguese texts, as described below.

\paragraph{Task Description} 
An annotation instance is an excerpt in Portuguese and its (machine) translation in English. Bilingual speakers are asked to read the two excerpts, determine whether the translation contains an accuracy error, and then rate its severity as being \textit{minor} or \textit{major}. 
Following \citet{freitag-etal-2021-experts}, minor errors are defined as errors that do not lead to loss of meaning and would not confuse or mislead the reader, while major ones may confuse or mislead the reader due to significant changes in meaning. We ask participants to factor out fluency issues in their assessment as much as possible.
We include a screenshot of the task description presented to annotators in Figures \ref{fig:instructions_2} and \ref{fig:interface_examples_2} of Appendix \ref{sec:appendix_interfaces}.

\paragraph{Conditions \& Evaluation Constructs} We use 
Precision, Recall, and F1 as our main evaluation measures and follow the same conditions and subjective measures described in \S\ref{sec:annotation}. The above scores are calculated only against major errors, as those are the ones for which we have gold labels. 
We report statistical significance based on bootstrap resampling as outlined in \S\ref{sec:annotation_measures}. For this study, which contains a total of $250$  annotations, the significance level is $p=0.05$.

\paragraph{Procedures \& Participants} We collect $5$ assessments per instance and annotate a total of $100$ instances ($50$ per condition). Each participant is presented with $25$ instances, among which $20\%$ correspond to \textsc{wmt} critical errors, $20\%$ to negation errors, and the rest to original \textsc{opus} parallel texts. We recruit a total of $40$ participants, all of whom self-identified as proficient in English and Portuguese. All other participant recruitment details and study procedures are the same as in \S\ref{sec:annotation}.

\begin{figure}[!t]
    \centering
    \scalebox{0.39}{
    \includegraphics{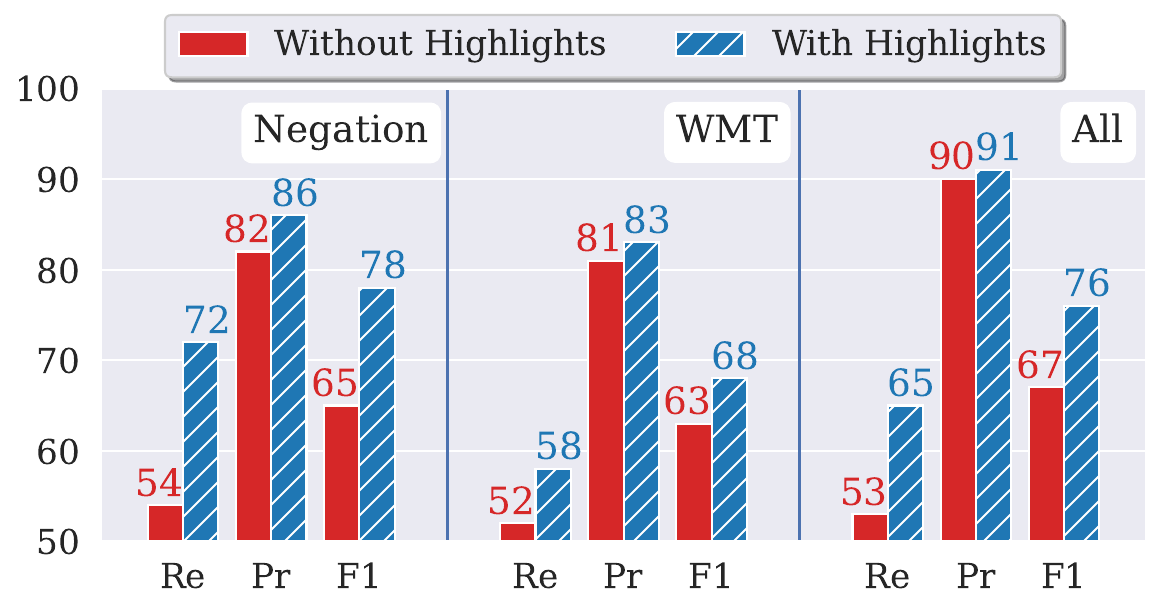}}
    \caption{
    Application-grounded evaluation results comparing \nh vs. \yh annotation conditions. Contrastive phrasal highlights significantly improve the Recall and F1 scores when detecting \textit{negation} errors ($p=0.05$) and errors in the \textsc{all} set. Precision improvements across sets and any improvements on the \textsc{wmt} set are not significant ($p>0.05$).
    }\label{fig:ced_accuracy}
\end{figure}

\subsection{Study Results}\label{sec:study_findings_2}

\paragraph{Main Findings} As shown in Figure~\ref{fig:ced_accuracy}, the user group presented with \hname highlights exhibits a higher recall in detecting critical errors compared to the one that does not access them. Crucially, we report significant improvements ($p=0.05$) in detecting \textit{negation} errors, which are harder to spot. 
A closer look at the differences between the two error classes reveals that detecting negation errors is more challenging as annotators have to pay close attention to the compared texts and the relationships between them to spot local mistranslations that cause a major shift in meaning. On the other hand, \textsc{wmt} errors are more easily spotted as they mostly represent detached hallucination phenomena (i.e., an entire phrase is added to the English translation), which we hypothesize can be implied by length differences of the compared texts. Our full annotation results, presented in Table~\ref{tab:ced_annotations}, further validate our hypothesis: \textsc{wmt} errors are not genuinely missed out but rather rated as minor by some annotators. On the other hand, usually, negation errors are not only misperceived as minor but are entirely overlooked. We include examples of annotations in Figure~\ref{fig:interface_examples_2}.

\begin{table}[!t]
    \centering
    \scalebox{0.35}{
    \begin{tabular}{ccccc@{\hskip 0.2in}ccccc@{\hskip 0.35in}cccccccccccccccc}

\rowcolor{gray!10}
\multicolumn{25}{c}{\yh}\\
\rowcolor{gray!10}
\multicolumn{5}{c}{Negation} & \multicolumn{5}{c}{\textsc{wmt}} & \multicolumn{15}{c}{No Error}\\
 \cellcolor{red!20} \major & \cellcolor{red!20} \major & \cellcolor{red!20} \major & \cellcolor{red!20} \major & \cellcolor{red!20} \major & \cellcolor{red!20} \major & \cellcolor{yellow!20} \minor & \cellcolor{red!20} \major & \cellcolor{red!20} \major & \cellcolor{red!20} \major & \cellcolor{green!20} \noerror & \cellcolor{green!20} \noerror & \cellcolor{green!20} \noerror & \cellcolor{green!20} \noerror & \cellcolor{green!20} \noerror & \cellcolor{green!20} \noerror & \cellcolor{green!20} \noerror & \cellcolor{green!20} \noerror & \cellcolor{green!20} \noerror & \cellcolor{green!20} \noerror & \cellcolor{green!20} \noerror & \cellcolor{green!20} \noerror & \cellcolor{green!20} \noerror & \cellcolor{green!20} \noerror & \cellcolor{yellow!20} \minor \\
\cellcolor{green!20} \noerror & \cellcolor{red!20} \major & \cellcolor{red!20} \major & \cellcolor{red!20} \major & \cellcolor{red!20} \major & \cellcolor{red!20} \major & \cellcolor{red!20} \major & \cellcolor{red!20} \major & \cellcolor{red!20} \major & \cellcolor{red!20} \major & \cellcolor{green!20} \noerror & \cellcolor{green!20} \noerror & \cellcolor{yellow!20} \minor & \cellcolor{green!20} \noerror & \cellcolor{green!20} \noerror & \cellcolor{green!20} \noerror & \cellcolor{green!20} \noerror & \cellcolor{green!20} \noerror & \cellcolor{yellow!20} \minor & \cellcolor{green!20} \noerror & \cellcolor{yellow!20} \minor & \cellcolor{yellow!20} \minor & \cellcolor{green!20} \noerror & \cellcolor{green!20} \noerror & \cellcolor{green!20} \noerror \\
\cellcolor{red!20} \major & \cellcolor{yellow!20} \minor & \cellcolor{red!20} \major & \cellcolor{green!20} \noerror & \cellcolor{red!20} \major & \cellcolor{red!20} \major & \cellcolor{red!20} \major & \cellcolor{red!20} \major & \cellcolor{red!20} \major & \cellcolor{red!20} \major & \cellcolor{green!20} \noerror & \cellcolor{green!20} \noerror & \cellcolor{green!20} \noerror & \cellcolor{green!20} \noerror & \cellcolor{green!20} \noerror & \cellcolor{red!20} \major & \cellcolor{green!20} \noerror & \cellcolor{yellow!20} \minor & \cellcolor{yellow!20} \minor & \cellcolor{yellow!20} \minor & \cellcolor{yellow!20} \minor & \cellcolor{green!20} \noerror & \cellcolor{yellow!20} \minor & \cellcolor{green!20} \noerror & \cellcolor{yellow!20} \minor \\
 \cellcolor{yellow!20} \minor & \cellcolor{red!20} \major & \cellcolor{red!20} \major & \cellcolor{red!20} \major & \cellcolor{red!20} \major & \cellcolor{red!20} \major & \cellcolor{yellow!20} \minor & \cellcolor{red!20} \major & \cellcolor{red!20} \major & \cellcolor{red!20} \major & \cellcolor{green!20} \noerror & \cellcolor{green!20} \noerror & \cellcolor{green!20} \noerror & \cellcolor{green!20} \noerror & \cellcolor{green!20} \noerror & \cellcolor{yellow!20} \minor & \cellcolor{green!20} \noerror & \cellcolor{green!20} \noerror & \cellcolor{green!20} \noerror & \cellcolor{green!20} \noerror & \cellcolor{green!20} \noerror & \cellcolor{red!20} \major & \cellcolor{green!20} \noerror & \cellcolor{green!20} \noerror & \cellcolor{red!20} \major \\
\cellcolor{red!20} \major & \cellcolor{red!20} \major & \cellcolor{red!20} \major & \cellcolor{red!20} \major & \cellcolor{red!20} \major & \cellcolor{red!20} \major & \cellcolor{yellow!20} \minor & \cellcolor{green!20} \noerror & \cellcolor{red!20} \major & \cellcolor{red!20} \major & \cellcolor{green!20} \noerror & \cellcolor{yellow!20} \minor & \cellcolor{green!20} \noerror & \cellcolor{yellow!20} \minor & \cellcolor{yellow!20} \minor & \cellcolor{green!20} \noerror & \cellcolor{green!20} \noerror & \cellcolor{green!20} \noerror & \cellcolor{green!20} \noerror & \cellcolor{green!20} \noerror & \cellcolor{green!20} \noerror & \cellcolor{green!20} \noerror & \cellcolor{green!20} \noerror & \cellcolor{green!20} \noerror & \cellcolor{yellow!20} \minor \\
\cellcolor{red!20} \major & \cellcolor{red!20} \major & \cellcolor{yellow!20} \minor & \cellcolor{red!20} \major & \cellcolor{red!20} \major & \cellcolor{red!20} \major & \cellcolor{red!20} \major & \cellcolor{red!20} \major & \cellcolor{red!20} \major & \cellcolor{green!20} \noerror & \cellcolor{green!20} \noerror & \cellcolor{green!20} \noerror & \cellcolor{green!20} \noerror & \cellcolor{red!20} \major & \cellcolor{green!20} \noerror & \cellcolor{green!20} \noerror & \cellcolor{green!20} \noerror & \cellcolor{green!20} \noerror & \cellcolor{green!20} \noerror & \cellcolor{green!20} \noerror & \cellcolor{yellow!20} \minor & \cellcolor{green!20} \noerror & \cellcolor{green!20} \noerror & \cellcolor{green!20} \noerror & \cellcolor{yellow!20} \minor \\
 \cellcolor{red!20} \major & \cellcolor{red!20} \major & \cellcolor{red!20} \major & \cellcolor{red!20} \major & \cellcolor{red!20} \major & \cellcolor{yellow!20} \minor & \cellcolor{yellow!20} \minor & \cellcolor{red!20} \major & \cellcolor{yellow!20} \minor & \cellcolor{red!20} \major & \cellcolor{green!20} \noerror & \cellcolor{green!20} \noerror & \cellcolor{green!20} \noerror & \cellcolor{green!20} \noerror & \cellcolor{green!20} \noerror & \cellcolor{green!20} \noerror & \cellcolor{green!20} \noerror & \cellcolor{green!20} \noerror & \cellcolor{green!20} \noerror & \cellcolor{green!20} \noerror & \cellcolor{yellow!20} \minor & \cellcolor{green!20} \noerror & \cellcolor{green!20} \noerror & \cellcolor{green!20} \noerror & \cellcolor{green!20} \noerror \\
\cellcolor{red!20} \major & \cellcolor{red!20} \major & \cellcolor{green!20} \noerror & \cellcolor{yellow!20} \minor & \cellcolor{red!20} \major & \cellcolor{red!20} \major & \cellcolor{yellow!20} \minor & \cellcolor{red!20} \major & \cellcolor{red!20} \major & \cellcolor{red!20} \major & \cellcolor{green!20} \noerror & \cellcolor{green!20} \noerror & \cellcolor{green!20} \noerror & \cellcolor{green!20} \noerror & \cellcolor{green!20} \noerror & \cellcolor{yellow!20} \minor & \cellcolor{yellow!20} \minor & \cellcolor{green!20} \noerror & \cellcolor{green!20} \noerror & \cellcolor{red!20} \major & \cellcolor{yellow!20} \minor & \cellcolor{green!20} \noerror & \cellcolor{green!20} \noerror & \cellcolor{green!20} \noerror & \cellcolor{red!20} \major \\
\cellcolor{red!20} \major & \cellcolor{red!20} \major & \cellcolor{green!20} \noerror & \cellcolor{red!20} \major & \cellcolor{red!20} \major & \cellcolor{red!20} \major & \cellcolor{yellow!20} \minor & \cellcolor{red!20} \major & \cellcolor{red!20} \major & \cellcolor{yellow!20} \minor & \cellcolor{green!20} \noerror & \cellcolor{green!20} \noerror & \cellcolor{green!20} \noerror & \cellcolor{red!20} \major & \cellcolor{green!20} \noerror & \cellcolor{yellow!20} \minor & \cellcolor{green!20} \noerror & \cellcolor{green!20} \noerror & \cellcolor{green!20} \noerror & \cellcolor{green!20} \noerror & \cellcolor{yellow!20} \minor & \cellcolor{green!20} \noerror & \cellcolor{green!20} \noerror & \cellcolor{green!20} \noerror & \cellcolor{yellow!20} \minor \\
\cellcolor{yellow!20} \minor & \cellcolor{red!20} \major & \cellcolor{red!20} \major & \cellcolor{red!20} \major & \cellcolor{red!20} \major & \cellcolor{red!20} \major & \cellcolor{yellow!20} \minor & \cellcolor{red!20} \major & \cellcolor{red!20} \major & \cellcolor{yellow!20} \minor & \cellcolor{green!20} \noerror & \cellcolor{green!20} \noerror & \cellcolor{green!20} \noerror & \cellcolor{green!20} \noerror & \cellcolor{green!20} \noerror & \cellcolor{green!20} \noerror & \cellcolor{green!20} \noerror & \cellcolor{green!20} \noerror & \cellcolor{green!20} \noerror & \cellcolor{green!20} \noerror & \cellcolor{green!20} \noerror & \cellcolor{yellow!20} \minor & \cellcolor{green!20} \noerror & \cellcolor{green!20} \noerror & \cellcolor{green!20} \noerror \\
\cellcolor{green!20} \noerror & \cellcolor{yellow!20} \minor & \cellcolor{red!20} \major & \cellcolor{red!20} \major & \cellcolor{red!20} \major & \cellcolor{yellow!20} \minor & \cellcolor{green!20} \noerror & \cellcolor{red!20} \major & \cellcolor{red!20} \major & \cellcolor{red!20} \major & \cellcolor{green!20} \noerror & \cellcolor{red!20} \major & \cellcolor{green!20} \noerror & \cellcolor{green!20} \noerror & \cellcolor{green!20} \noerror & \cellcolor{green!20} \noerror & \cellcolor{green!20} \noerror & \cellcolor{green!20} \noerror & \cellcolor{green!20} \noerror & \cellcolor{green!20} \noerror & \cellcolor{red!20} \major & \cellcolor{green!20} \noerror & \cellcolor{yellow!20} \minor & \cellcolor{green!20} \noerror & \cellcolor{green!20} \noerror \\
\cellcolor{red!20} \major & \cellcolor{red!20} \major & \cellcolor{red!20} \major & \cellcolor{yellow!20} \minor & \cellcolor{red!20} \major & \cellcolor{yellow!20} \minor & \cellcolor{yellow!20} \minor & \cellcolor{red!20} \major & \cellcolor{yellow!20} \minor & \cellcolor{red!20} \major & \cellcolor{green!20} \noerror & \cellcolor{green!20} \noerror & \cellcolor{green!20} \noerror & \cellcolor{green!20} \noerror & \cellcolor{green!20} \noerror & \cellcolor{green!20} \noerror & \cellcolor{green!20} \noerror & \cellcolor{green!20} \noerror & \cellcolor{green!20} \noerror & \cellcolor{green!20} \noerror & \cellcolor{green!20} \noerror & \cellcolor{green!20} \noerror & \cellcolor{green!20} \noerror & \cellcolor{green!20} \noerror & \cellcolor{green!20} \noerror \\
\cellcolor{yellow!20} \minor & \cellcolor{red!20} \major & \cellcolor{green!20} \noerror & \cellcolor{red!20} \major & \cellcolor{yellow!20} \minor & \cellcolor{red!20} \major & \cellcolor{yellow!20} \minor & \cellcolor{red!20} \major & \cellcolor{red!20} \major & \cellcolor{green!20} \noerror & \cellcolor{green!20} \noerror & \cellcolor{green!20} \noerror & \cellcolor{green!20} \noerror & \cellcolor{yellow!20} \minor & \cellcolor{red!20} \major & \cellcolor{yellow!20} \minor & \cellcolor{green!20} \noerror & \cellcolor{green!20} \noerror & \cellcolor{green!20} \noerror & \cellcolor{yellow!20} \minor & \cellcolor{green!20} \noerror & \cellcolor{green!20} \noerror & \cellcolor{green!20} \noerror & \cellcolor{green!20} \noerror & \cellcolor{green!20} \noerror \\
\cellcolor{red!20} \major & \cellcolor{red!20} \major & \cellcolor{green!20} \noerror & \cellcolor{red!20} \major & \cellcolor{yellow!20} \minor & \cellcolor{red!20} \major & \cellcolor{green!20} \noerror & \cellcolor{yellow!20} \minor & \cellcolor{red!20} \major & \cellcolor{yellow!20} \minor & \cellcolor{green!20} \noerror & \cellcolor{green!20} \noerror & \cellcolor{green!20} \noerror & \cellcolor{red!20} \major & \cellcolor{green!20} \noerror & \cellcolor{green!20} \noerror & \cellcolor{green!20} \noerror & \cellcolor{yellow!20} \minor & \cellcolor{green!20} \noerror & \cellcolor{yellow!20} \minor & \cellcolor{green!20} \noerror & \cellcolor{yellow!20} \minor & \cellcolor{green!20} \noerror & \cellcolor{green!20} \noerror & \cellcolor{yellow!20} \minor \\
\cellcolor{yellow!20} \minor & \cellcolor{red!20} \major & \cellcolor{yellow!20} \minor & \cellcolor{red!20} \major & \cellcolor{red!20} \major & \cellcolor{red!20} \major & \cellcolor{yellow!20} \minor & \cellcolor{yellow!20} \minor & \cellcolor{red!20} \major & \cellcolor{green!20} \noerror & \cellcolor{green!20} \noerror & \cellcolor{green!20} \noerror & \cellcolor{green!20} \noerror & \cellcolor{yellow!20} \minor & \cellcolor{green!20} \noerror & \cellcolor{green!20} \noerror & \cellcolor{green!20} \noerror & \cellcolor{green!20} \noerror & \cellcolor{green!20} \noerror & \cellcolor{yellow!20} \minor & \cellcolor{green!20} \noerror & \cellcolor{green!20} \noerror & \cellcolor{green!20} \noerror & \cellcolor{green!20} \noerror & \cellcolor{red!20} \major \\
\cellcolor{red!20} \major & \cellcolor{red!20} \major & \cellcolor{red!20} \major & \cellcolor{yellow!20} \minor & \cellcolor{red!20} \major & \cellcolor{yellow!20} \minor & \cellcolor{yellow!20} \minor & \cellcolor{yellow!20} \minor & \cellcolor{yellow!20} \minor & \cellcolor{red!20} \major & \cellcolor{green!20} \noerror & \cellcolor{green!20} \noerror & \cellcolor{yellow!20} \minor & \cellcolor{green!20} \noerror & \cellcolor{green!20} \noerror & \cellcolor{red!20} \major & \cellcolor{green!20} \noerror & \cellcolor{green!20} \noerror & \cellcolor{green!20} \noerror & \cellcolor{yellow!20} \minor & \cellcolor{green!20} \noerror & \cellcolor{green!20} \noerror & \cellcolor{green!20} \noerror & \cellcolor{green!20} \noerror & \cellcolor{green!20} \noerror \\
\cellcolor{green!20} \noerror & \cellcolor{red!20} \major & \cellcolor{red!20} \major & \cellcolor{green!20} \noerror & \cellcolor{red!20} \major & \cellcolor{yellow!20} \minor & \cellcolor{yellow!20} \minor & \cellcolor{red!20} \major & \cellcolor{yellow!20} \minor & \cellcolor{red!20} \major & \cellcolor{green!20} \noerror & \cellcolor{green!20} \noerror & \cellcolor{green!20} \noerror & \cellcolor{green!20} \noerror & \cellcolor{green!20} \noerror & \cellcolor{yellow!20} \minor & \cellcolor{green!20} \noerror & \cellcolor{green!20} \noerror & \cellcolor{green!20} \noerror & \cellcolor{green!20} \noerror & \cellcolor{green!20} \noerror & \cellcolor{green!20} \noerror & \cellcolor{green!20} \noerror & \cellcolor{green!20} \noerror & \cellcolor{yellow!20} \minor \\
\cellcolor{red!20} \major & \cellcolor{yellow!20} \minor & \cellcolor{red!20} \major & \cellcolor{yellow!20} \minor & \cellcolor{red!20} \major & \cellcolor{yellow!20} \minor & \cellcolor{yellow!20} \minor & \cellcolor{red!20} \major & \cellcolor{yellow!20} \minor & \cellcolor{red!20} \major & \cellcolor{green!20} \noerror & \cellcolor{green!20} \noerror & \cellcolor{yellow!20} \minor & \cellcolor{green!20} \noerror & \cellcolor{green!20} \noerror & \cellcolor{yellow!20} \minor & \cellcolor{green!20} \noerror & \cellcolor{green!20} \noerror & \cellcolor{green!20} \noerror & \cellcolor{green!20} \noerror & \cellcolor{yellow!20} \minor & \cellcolor{green!20} \noerror & \cellcolor{yellow!20} \minor & \cellcolor{yellow!20} \minor & \cellcolor{green!20} \noerror \\
\cellcolor{red!20} \major & \cellcolor{green!20} \noerror & \cellcolor{green!20} \noerror & \cellcolor{green!20} \noerror & \cellcolor{red!20} \major & \cellcolor{green!20} \noerror & \cellcolor{red!20} \major & \cellcolor{yellow!20} \minor & \cellcolor{red!20} \major & \cellcolor{green!20} \noerror & \cellcolor{green!20} \noerror & \cellcolor{green!20} \noerror & \cellcolor{green!20} \noerror & \cellcolor{green!20} \noerror & \cellcolor{green!20} \noerror & \cellcolor{red!20} \major & \cellcolor{red!20} \major & \cellcolor{green!20} \noerror & \cellcolor{green!20} \noerror & \cellcolor{yellow!20} \minor & \cellcolor{yellow!20} \minor & \cellcolor{green!20} \noerror & \cellcolor{yellow!20} \minor & \cellcolor{green!20} \noerror & \cellcolor{green!20} \noerror \\
\cellcolor{yellow!20} \minor & \cellcolor{yellow!20} \minor & \cellcolor{red!20} \major & \cellcolor{red!20} \major & \cellcolor{red!20} \major & \cellcolor{yellow!20} \minor & \cellcolor{yellow!20} \minor & \cellcolor{yellow!20} \minor & \cellcolor{yellow!20} \minor & \cellcolor{red!20} \major & \cellcolor{green!20} \noerror & \cellcolor{green!20} \noerror & \cellcolor{green!20} \noerror & \cellcolor{green!20} \noerror & \cellcolor{green!20} \noerror & \cellcolor{yellow!20} \minor & \cellcolor{green!20} \noerror & \cellcolor{green!20} \noerror & \cellcolor{green!20} \noerror & \cellcolor{yellow!20} \minor & \cellcolor{green!20} \noerror & \cellcolor{green!20} \noerror & \cellcolor{green!20} \noerror & \cellcolor{green!20} \noerror & \cellcolor{green!20} \noerror \\
\\
\rowcolor{gray!10}
\multicolumn{25}{c}{\nh}\\
\rowcolor{gray!10}
\multicolumn{5}{c}{Negation} & \multicolumn{5}{c}{\textsc{wmt}} & \multicolumn{15}{c}{No Error}\\
\cellcolor{red!20} \major & \cellcolor{red!20} \major & \cellcolor{red!20} \major & \cellcolor{red!20} \major & \cellcolor{red!20} \major & \cellcolor{red!20} \major & \cellcolor{yellow!20} \minor & \cellcolor{red!20} \major & \cellcolor{red!20} \major & \cellcolor{red!20} \major & \cellcolor{green!20} \noerror & \cellcolor{green!20} \noerror & \cellcolor{green!20} \noerror & \cellcolor{green!20} \noerror & \cellcolor{green!20} \noerror & \cellcolor{green!20} \noerror & \cellcolor{green!20} \noerror & \cellcolor{green!20} \noerror & \cellcolor{green!20} \noerror & \cellcolor{green!20} \noerror & \cellcolor{green!20} \noerror & \cellcolor{green!20} \noerror & \cellcolor{green!20} \noerror & \cellcolor{green!20} \noerror & \cellcolor{red!20} \major \\
\cellcolor{red!20} \major & \cellcolor{red!20} \major & \cellcolor{red!20} \major & \cellcolor{red!20} \major & \cellcolor{yellow!20} \minor & \cellcolor{red!20} \major & \cellcolor{red!20} \major & \cellcolor{red!20} \major & \cellcolor{red!20} \major & \cellcolor{yellow!20} \minor & \cellcolor{yellow!20} \minor & \cellcolor{yellow!20} \minor & \cellcolor{yellow!20} \minor & \cellcolor{yellow!20} \minor & \cellcolor{yellow!20} \minor & \cellcolor{yellow!20} \minor & \cellcolor{yellow!20} \minor & \cellcolor{yellow!20} \minor & \cellcolor{green!20} \noerror & \cellcolor{yellow!20} \minor & \cellcolor{yellow!20} \minor & \cellcolor{yellow!20} \minor & \cellcolor{yellow!20} \minor & \cellcolor{yellow!20} \minor & \cellcolor{yellow!20} \minor \\
\cellcolor{green!20} \noerror & \cellcolor{red!20} \major & \cellcolor{red!20} \major & \cellcolor{red!20} \major & \cellcolor{red!20} \major & \cellcolor{red!20} \major & \cellcolor{red!20} \major & \cellcolor{red!20} \major & \cellcolor{red!20} \major & \cellcolor{green!20} \noerror & \cellcolor{green!20} \noerror & \cellcolor{green!20} \noerror & \cellcolor{green!20} \noerror & \cellcolor{green!20} \noerror & \cellcolor{green!20} \noerror & \cellcolor{green!20} \noerror & \cellcolor{green!20} \noerror & \cellcolor{green!20} \noerror & \cellcolor{green!20} \noerror & \cellcolor{red!20} \major & \cellcolor{red!20} \major & \cellcolor{green!20} \noerror & \cellcolor{green!20} \noerror & \cellcolor{green!20} \noerror & \cellcolor{red!20} \major \\
 \cellcolor{red!20} \major & \cellcolor{green!20} \noerror & \cellcolor{red!20} \major & \cellcolor{red!20} \major & \cellcolor{yellow!20} \minor & \cellcolor{red!20} \major & \cellcolor{red!20} \major & \cellcolor{red!20} \major & \cellcolor{red!20} \major & \cellcolor{red!20} \major & \cellcolor{green!20} \noerror & \cellcolor{green!20} \noerror & \cellcolor{red!20} \major & \cellcolor{yellow!20} \minor & \cellcolor{green!20} \noerror & \cellcolor{yellow!20} \minor & \cellcolor{green!20} \noerror & \cellcolor{green!20} \noerror & \cellcolor{yellow!20} \minor & \cellcolor{green!20} \noerror & \cellcolor{red!20} \major & \cellcolor{green!20} \noerror & \cellcolor{green!20} \noerror & \cellcolor{green!20} \noerror & \cellcolor{green!20} \noerror \\
 \cellcolor{green!20} \noerror & \cellcolor{red!20} \major & \cellcolor{red!20} \major & \cellcolor{yellow!20} \minor & \cellcolor{red!20} \major & \cellcolor{red!20} \major & \cellcolor{red!20} \major & \cellcolor{red!20} \major & \cellcolor{red!20} \major & \cellcolor{yellow!20} \minor & \cellcolor{green!20} \noerror & \cellcolor{green!20} \noerror & \cellcolor{green!20} \noerror & \cellcolor{green!20} \noerror & \cellcolor{yellow!20} \minor & \cellcolor{green!20} \noerror & \cellcolor{yellow!20} \minor & \cellcolor{yellow!20} \minor & \cellcolor{green!20} \noerror & \cellcolor{green!20} \noerror & \cellcolor{green!20} \noerror & \cellcolor{green!20} \noerror & \cellcolor{green!20} \noerror & \cellcolor{green!20} \noerror & \cellcolor{yellow!20} \minor \\
 \cellcolor{green!20} \noerror & \cellcolor{red!20} \major & \cellcolor{red!20} \major & \cellcolor{red!20} \major & \cellcolor{red!20} \major & \cellcolor{red!20} \major & \cellcolor{yellow!20} \minor & \cellcolor{green!20} \noerror & \cellcolor{red!20} \major & \cellcolor{red!20} \major & \cellcolor{green!20} \noerror & \cellcolor{green!20} \noerror & \cellcolor{green!20} \noerror & \cellcolor{green!20} \noerror & \cellcolor{green!20} \noerror & \cellcolor{green!20} \noerror & \cellcolor{yellow!20} \minor & \cellcolor{green!20} \noerror & \cellcolor{green!20} \noerror & \cellcolor{green!20} \noerror & \cellcolor{green!20} \noerror & \cellcolor{green!20} \noerror & \cellcolor{green!20} \noerror & \cellcolor{green!20} \noerror & \cellcolor{red!20} \major \\
\cellcolor{red!20} \major & \cellcolor{red!20} \major & \cellcolor{red!20} \major & \cellcolor{red!20} \major & \cellcolor{red!20} \major & \cellcolor{red!20} \major & \cellcolor{yellow!20} \minor & \cellcolor{yellow!20} \minor & \cellcolor{red!20} \major & \cellcolor{yellow!20} \minor & \cellcolor{yellow!20} \minor & \cellcolor{green!20} \noerror & \cellcolor{green!20} \noerror & \cellcolor{green!20} \noerror & \cellcolor{green!20} \noerror & \cellcolor{green!20} \noerror & \cellcolor{yellow!20} \minor & \cellcolor{green!20} \noerror & \cellcolor{green!20} \noerror & \cellcolor{green!20} \noerror & \cellcolor{green!20} \noerror & \cellcolor{yellow!20} \minor & \cellcolor{green!20} \noerror & \cellcolor{green!20} \noerror & \cellcolor{yellow!20} \minor \\
\cellcolor{green!20} \noerror & \cellcolor{green!20} \noerror & \cellcolor{red!20} \major & \cellcolor{red!20} \major & \cellcolor{red!20} \major & \cellcolor{yellow!20} \minor & \cellcolor{red!20} \major & \cellcolor{red!20} \major & \cellcolor{red!20} \major & \cellcolor{red!20} \major & \cellcolor{yellow!20} \minor & \cellcolor{green!20} \noerror & \cellcolor{green!20} \noerror & \cellcolor{green!20} \noerror & \cellcolor{green!20} \noerror & \cellcolor{yellow!20} \minor & \cellcolor{green!20} \noerror & \cellcolor{green!20} \noerror & \cellcolor{yellow!20} \minor & \cellcolor{yellow!20} \minor & \cellcolor{yellow!20} \minor & \cellcolor{green!20} \noerror & \cellcolor{green!20} \noerror & \cellcolor{yellow!20} \minor & \cellcolor{yellow!20} \minor \\
 \cellcolor{yellow!20} \minor & \cellcolor{yellow!20} \minor & \cellcolor{red!20} \major & \cellcolor{red!20} \major & \cellcolor{red!20} \major & \cellcolor{green!20} \noerror & \cellcolor{red!20} \major & \cellcolor{red!20} \major & \cellcolor{green!20} \noerror & \cellcolor{red!20} \major & \cellcolor{green!20} \noerror & \cellcolor{green!20} \noerror & \cellcolor{green!20} \noerror & \cellcolor{green!20} \noerror & \cellcolor{green!20} \noerror & \cellcolor{green!20} \noerror & \cellcolor{green!20} \noerror & \cellcolor{green!20} \noerror & \cellcolor{green!20} \noerror & \cellcolor{green!20} \noerror & \cellcolor{green!20} \noerror & \cellcolor{green!20} \noerror & \cellcolor{green!20} \noerror & \cellcolor{green!20} \noerror & \cellcolor{green!20} \noerror \\
\cellcolor{red!20} \major & \cellcolor{red!20} \major & \cellcolor{red!20} \major & \cellcolor{red!20} \major & \cellcolor{yellow!20} \minor & \cellcolor{green!20} \noerror & \cellcolor{green!20} \noerror & \cellcolor{red!20} \major & \cellcolor{green!20} \noerror & \cellcolor{red!20} \major & \cellcolor{green!20} \noerror & \cellcolor{green!20} \noerror & \cellcolor{yellow!20} \minor & \cellcolor{green!20} \noerror & \cellcolor{green!20} \noerror & \cellcolor{red!20} \major & \cellcolor{green!20} \noerror & \cellcolor{green!20} \noerror & \cellcolor{yellow!20} \minor & \cellcolor{green!20} \noerror & \cellcolor{green!20} \noerror & \cellcolor{green!20} \noerror & \cellcolor{yellow!20} \minor & \cellcolor{green!20} \noerror & \cellcolor{green!20} \noerror \\
\cellcolor{green!20} \noerror & \cellcolor{yellow!20} \minor & \cellcolor{red!20} \major & \cellcolor{red!20} \major & \cellcolor{red!20} \major & \cellcolor{red!20} \major & \cellcolor{yellow!20} \minor & \cellcolor{red!20} \major & \cellcolor{red!20} \major & \cellcolor{green!20} \noerror & \cellcolor{green!20} \noerror & \cellcolor{green!20} \noerror & \cellcolor{green!20} \noerror & \cellcolor{green!20} \noerror & \cellcolor{yellow!20} \minor & \cellcolor{yellow!20} \minor & \cellcolor{yellow!20} \minor & \cellcolor{yellow!20} \minor & \cellcolor{green!20} \noerror & \cellcolor{green!20} \noerror & \cellcolor{green!20} \noerror & \cellcolor{green!20} \noerror & \cellcolor{green!20} \noerror & \cellcolor{green!20} \noerror & \cellcolor{green!20} \noerror \\
 \cellcolor{red!20} \major & \cellcolor{yellow!20} \minor & \cellcolor{red!20} \major & \cellcolor{red!20} \major & \cellcolor{red!20} \major & \cellcolor{yellow!20} \minor & \cellcolor{yellow!20} \minor & \cellcolor{yellow!20} \minor & \cellcolor{red!20} \major & \cellcolor{green!20} \noerror & \cellcolor{green!20} \noerror & \cellcolor{green!20} \noerror & \cellcolor{green!20} \noerror & \cellcolor{yellow!20} \minor & \cellcolor{yellow!20} \minor & \cellcolor{green!20} \noerror & \cellcolor{green!20} \noerror & \cellcolor{green!20} \noerror & \cellcolor{green!20} \noerror & \cellcolor{green!20} \noerror & \cellcolor{yellow!20} \minor & \cellcolor{green!20} \noerror & \cellcolor{green!20} \noerror & \cellcolor{green!20} \noerror & \cellcolor{green!20} \noerror \\
 \cellcolor{green!20} \noerror & \cellcolor{yellow!20} \minor & \cellcolor{green!20} \noerror & \cellcolor{green!20} \noerror & \cellcolor{red!20} \major & \cellcolor{red!20} \major & \cellcolor{green!20} \noerror & \cellcolor{yellow!20} \minor & \cellcolor{red!20} \major & \cellcolor{red!20} \major & \cellcolor{green!20} \noerror & \cellcolor{green!20} \noerror & \cellcolor{green!20} \noerror & \cellcolor{green!20} \noerror & \cellcolor{green!20} \noerror & \cellcolor{green!20} \noerror & \cellcolor{green!20} \noerror & \cellcolor{green!20} \noerror & \cellcolor{green!20} \noerror & \cellcolor{green!20} \noerror & \cellcolor{green!20} \noerror & \cellcolor{green!20} \noerror & \cellcolor{green!20} \noerror & \cellcolor{green!20} \noerror & \cellcolor{green!20} \noerror \\
 \cellcolor{green!20} \noerror & \cellcolor{green!20} \noerror & \cellcolor{red!20} \major & \cellcolor{red!20} \major & \cellcolor{green!20} \noerror & \cellcolor{yellow!20} \minor & \cellcolor{yellow!20} \minor & \cellcolor{red!20} \major & \cellcolor{yellow!20} \minor & \cellcolor{red!20} \major & \cellcolor{yellow!20} \minor & \cellcolor{green!20} \noerror & \cellcolor{green!20} \noerror & \cellcolor{green!20} \noerror & \cellcolor{green!20} \noerror & \cellcolor{green!20} \noerror & \cellcolor{green!20} \noerror & \cellcolor{green!20} \noerror & \cellcolor{yellow!20} \minor & \cellcolor{green!20} \noerror & \cellcolor{yellow!20} \minor & \cellcolor{green!20} \noerror & \cellcolor{green!20} \noerror & \cellcolor{yellow!20} \minor & \cellcolor{green!20} \noerror \\
 \cellcolor{green!20} \noerror & \cellcolor{green!20} \noerror & \cellcolor{red!20} \major & \cellcolor{green!20} \noerror & \cellcolor{green!20} \noerror & \cellcolor{green!20} \noerror & \cellcolor{red!20} \major & \cellcolor{red!20} \major & \cellcolor{red!20} \major & \cellcolor{green!20} \noerror & \cellcolor{green!20} \noerror & \cellcolor{green!20} \noerror & \cellcolor{green!20} \noerror & \cellcolor{green!20} \noerror & \cellcolor{green!20} \noerror & \cellcolor{green!20} \noerror & \cellcolor{green!20} \noerror & \cellcolor{green!20} \noerror & \cellcolor{yellow!20} \minor & \cellcolor{green!20} \noerror & \cellcolor{yellow!20} \minor & \cellcolor{green!20} \noerror & \cellcolor{green!20} \noerror & \cellcolor{green!20} \noerror & \cellcolor{red!20} \major \\
 \cellcolor{green!20} \noerror & \cellcolor{green!20} \noerror & \cellcolor{yellow!20} \minor & \cellcolor{red!20} \major & \cellcolor{green!20} \noerror & \cellcolor{green!20} \noerror & \cellcolor{yellow!20} \minor & \cellcolor{red!20} \major & \cellcolor{green!20} \noerror & \cellcolor{red!20} \major & \cellcolor{green!20} \noerror & \cellcolor{green!20} \noerror & \cellcolor{green!20} \noerror & \cellcolor{green!20} \noerror & \cellcolor{green!20} \noerror & \cellcolor{green!20} \noerror & \cellcolor{green!20} \noerror & \cellcolor{green!20} \noerror & \cellcolor{yellow!20} \minor & \cellcolor{green!20} \noerror & \cellcolor{green!20} \noerror & \cellcolor{green!20} \noerror & \cellcolor{yellow!20} \minor & \cellcolor{green!20} \noerror & \cellcolor{green!20} \noerror \\
\cellcolor{green!20} \noerror & \cellcolor{green!20} \noerror & \cellcolor{yellow!20} \minor & \cellcolor{red!20} \major & \cellcolor{green!20} \noerror & \cellcolor{green!20} \noerror & \cellcolor{yellow!20} \minor & \cellcolor{red!20} \major & \cellcolor{green!20} \noerror & \cellcolor{red!20} \major & \cellcolor{green!20} \noerror & \cellcolor{green!20} \noerror & \cellcolor{green!20} \noerror & \cellcolor{green!20} \noerror & \cellcolor{green!20} \noerror & \cellcolor{yellow!20} \minor & \cellcolor{green!20} \noerror & \cellcolor{green!20} \noerror & \cellcolor{yellow!20} \minor & \cellcolor{green!20} \noerror & \cellcolor{green!20} \noerror & \cellcolor{green!20} \noerror & \cellcolor{yellow!20} \minor & \cellcolor{green!20} \noerror & \cellcolor{green!20} \noerror \\

 \cellcolor{red!20} \major & \cellcolor{green!20} \noerror & \cellcolor{red!20} \major & \cellcolor{green!20} \noerror & \cellcolor{green!20} \noerror & \cellcolor{yellow!20} \minor & \cellcolor{yellow!20} \minor & \cellcolor{red!20} \major & \cellcolor{yellow!20} \minor & \cellcolor{yellow!20} \minor & \cellcolor{green!20} \noerror & \cellcolor{green!20} \noerror & \cellcolor{yellow!20} \minor & \cellcolor{green!20} \noerror & \cellcolor{green!20} \noerror & \cellcolor{green!20} \noerror & \cellcolor{green!20} \noerror & \cellcolor{yellow!20} \minor & \cellcolor{yellow!20} \minor & \cellcolor{green!20} \noerror & \cellcolor{yellow!20} \minor & \cellcolor{green!20} \noerror & \cellcolor{green!20} \noerror & \cellcolor{green!20} \noerror & \cellcolor{green!20} \noerror \\
\cellcolor{yellow!20} \minor & \cellcolor{red!20} \major & \cellcolor{green!20} \noerror & \cellcolor{yellow!20} \minor & \cellcolor{red!20} \major & \cellcolor{yellow!20} \minor & \cellcolor{yellow!20} \minor & \cellcolor{green!20} \noerror & \cellcolor{yellow!20} \minor & \cellcolor{green!20} \noerror & \cellcolor{red!20} \major & \cellcolor{green!20} \noerror & \cellcolor{green!20} \noerror & \cellcolor{yellow!20} \minor & \cellcolor{yellow!20} \minor & \cellcolor{red!20} \major & \cellcolor{red!20} \major & \cellcolor{yellow!20} \minor & \cellcolor{red!20} \major & \cellcolor{green!20} \noerror & \cellcolor{yellow!20} \minor & \cellcolor{green!20} \noerror & \cellcolor{yellow!20} \minor & \cellcolor{green!20} \noerror & \cellcolor{green!20} \noerror \\
 \cellcolor{green!20} \noerror & \cellcolor{green!20} \noerror & \cellcolor{green!20} \noerror & \cellcolor{green!20} \noerror & \cellcolor{green!20} \noerror & \cellcolor{red!20} \major & \cellcolor{yellow!20} \minor & \cellcolor{yellow!20} \minor & \cellcolor{yellow!20} \minor & \cellcolor{red!20} \major & \cellcolor{green!20} \noerror & \cellcolor{green!20} \noerror & \cellcolor{green!20} \noerror & \cellcolor{green!20} \noerror & \cellcolor{green!20} \noerror & \cellcolor{green!20} \noerror & \cellcolor{green!20} \noerror & \cellcolor{green!20} \noerror & \cellcolor{green!20} \noerror & \cellcolor{green!20} \noerror & \cellcolor{red!20} \major & \cellcolor{green!20} \noerror & \cellcolor{yellow!20} \minor & \cellcolor{green!20} \noerror & \cellcolor{green!20} \noerror \\

    \end{tabular}}
    \caption{Critical Error Detection user study results. Each row represents an annotator, and each cell is an annotation, where \noerror \ indicates no accuracy error detected, \minor \ indicates a minor error, and \major \ a critical error. }
    \label{tab:ced_annotations}
\end{table}


\paragraph{Subjective Measures \& User Feedback} 
Overall, bilingual speakers found the highlights helpful, with an average self-reported usefulness of $3.9$. Additionally, they reported they would like to use them as a tool to assist them in detecting critical errors, with an average score of $3.8$. 
We include details on the distribution of ratings in Table~\ref{tab:application_measures_ced}. 
Last, a closer look at the users' feedback sheds some light on our approach's current strengths and limitations. Although highlights were in principle useful and a \textit{``good feature to have especially when proofreading a great amount of translated texts''}, some annotators spotted \textit{``a high percentage of false positives''} (see Figures \ref{fig:ced_na_1} and \ref{fig:ced_na_2} in Appendix~\ref{sec:appendix_interfaces})
and raised concerns that \textit{``relying solely on them might make them less aware of errors hidden in non-highlighted texts''} (see Figure \ref{fig:ced_negation_2} in Appendix~\ref{sec:appendix_interfaces}).

\section{Conclusion}

We introduce an approach to extracting \hname highlights that explain \textsc{nlp} models that take as input \textit{two} texts.
Unlike existing techniques that highlight input \textit{tokens} independently, we consider the input's structure and explicitly model the relationships between contrasting inputs. 

We study the effectiveness of \hname highlights by explaining the predictions of a divergence ranker that compares and contrasts meaning differences in bilingual texts. Our proxy evaluation confirms 
our approach outperforms standard explainability approaches, that highlight tokens, by matching human-provided rationales of divergences in English-French Wikipedia texts. Finally, through a series of human-centered evaluations, we explore the usefulness of \hname highlights in application-grounded contexts for three language pairs. 
Our results suggest that \hname highlights assist bilingual speakers in detecting fine-grained meaning differences in (human) translated texts and critical errors due to local mistranslations in machine-translated texts. 

Our findings create opportunities for designing better machine-in-the-loop pipelines to
identify critical machine translation errors grounded in high-stake settings, study translation data at scale, and facilitate the creation of multilingual content through crowd-based efforts.

\section*{Acknowledgements}

We thank Sweta Agrawal, Hal Daumé III, Luke Zettlemoyer, Philip Resnik, Leo Zhicheng Liu, the anonymous reviewers, and the \textsc{clip} lab at \textsc{umd} for helpful comments. 
This work was funded in part by the U.S. Army Grant No. W911NF2120076 and by the National Science Foundation under Awards No. 1750695 and 2147292.
Any opinions, findings, and conclusions or recommendations expressed in this material are those of the authors.

\section*{Limitations}

Our work contributed evidence that contrastive phrasal highlights can provide a framework that assists humans in detecting meaning differences of a specific nature (i.e., fine-grained translation processes or local critical errors). However, detecting translation differences in the wild requires covering the entire distribution of meaning differences~\cite{freitag-etal-2021-experts}, which we leave for future work. 

Moreover, although our method builds on unsupervised modules (i.e., Divergent m\textsc{bert} and SimAlign)  that do not rely on supervised data and can, in principle, be applied to any language pair, we have only evaluated our approach in high-resource settings where we expect both the explanandum and the word alignment models to be reasonably accurate.
Therefore, further work should be conducted to explore how our findings generalize to other settings, such as in low-resource regimes where we might expect the alignment to be of poorer quality, introducing errors that may impact humans' perceived understanding of the contrastive highlights differently. 

Additionally, future work should explore how the alignment module should be operationalized for other \textsc{nlp} tasks that take as input two texts, potentially exploring monolingual aligners~\cite{lan-etal-2021-neural} or additional structured information such as abstract meaning representations~\cite{banarescu-etal-2013-abstract}. 

Finally, the significance of our results could be strengthened by increasing the sample sizes of future user studies. We view our current findings as solid starting points for an in-depth exploration of the usefulness of highlights for human-centered applications that future studies can build upon and extend with broad explorations.

\section*{Ethics Statement}

Both studies are approved by the University of Maryland Institutional Review Board (\textsc{irb} number $2018458-1$). As discussed in the paper, all participants involved in our studies gave their consent and were compensated at a rate of $15$ \textsc{usd} per hour. We collected minimal demographic information, and participants could opt out of answering the demographic questions. We did not collect any further potentially identifiable information. Participation in our study was completely voluntary, and participants had the right to withdraw at any time.

\bibliography{anthology,custom}
\bibliographystyle{acl_natbib}

\appendix
\section{Appendices}

\subsection{Explanandum Training Details}\label{sec:appendix_explanandum_details}
The divergence ranking model is trained using the public implementation of Divergent m\textsc{bert}~\cite{briakou-carpuat-2020-detecting}. Synthetic divergences are generated starting from the $5{,}000$ top-scoring WikiMatrix sentences based on \textsc{laser} score (i.e., seed equivalents). We fine-tune the “BERT-Base Multilingual Cased” model~\cite{devlin-etal-2019-bert} and set the margin equal to 5 as per the original implementation.

\subsection{Subjective Measures}
\begin{table}[!ht]
    \centering
    \scalebox{0.7}{
    \begin{tabular}{lll}
    \emojiagreemost    &\datavizha{7} &   \datavizha{4}\\
     \emojiagree    & \dataviza{6} &  \dataviza{12}\\
    \emojineutral    & \datavizn{4} &   \datavizn{2}\\
    \emojidisagree    & \datavizd{3} & \datavizd{0}\\
    \emojidisagreemost   & \datavizhd{0} & \datavizhd{2}\\
    \end{tabular}}
    \caption{Bilingual speakers' agreement ratings with the statement ``Highlights were useful in helping me detect meaning differences in the compared texts" (left)  and ``I would like to use the highlights to help me detect critical errors" (right).}
    \label{tab:application_measures_ced}
\end{table}

\subsection{ChatGPT Details}\label{sec:chatGPT_details}

Starting from a given English sentence, we asked Chat\textsc{gpt} to edit the sentence in a way that introduces a small meaning difference. Below, we include the different prompts we used:

\begin{itemize}
    \item \textit{``Can you edit a small phrase in sentence X to introduce a small meaning difference?''}
    \item  \textit{``Can you delete/add a phrase in sentence X to introduce a small difference in meaning?''}
    \item \textit{``Can you edit a small phrase that makes sentence X more general/explicit?''}
\end{itemize}

We then reviewed the edits introduced by Chat\textsc{gpt}to make sure it indeed introduced small meaning differences. If we are not satisfied by a current edit, we ask again and potentially specify what parts of the sentence we would like to be edited.

\subsection{User Study Interfaces}\label{sec:appendix_interfaces} 
Below we include screenshots of our user study interfaces along with annotation examples for each of our evaluation tasks.

\begin{figure*}[h]
    \centering
    \includegraphics[width=1.\linewidth]{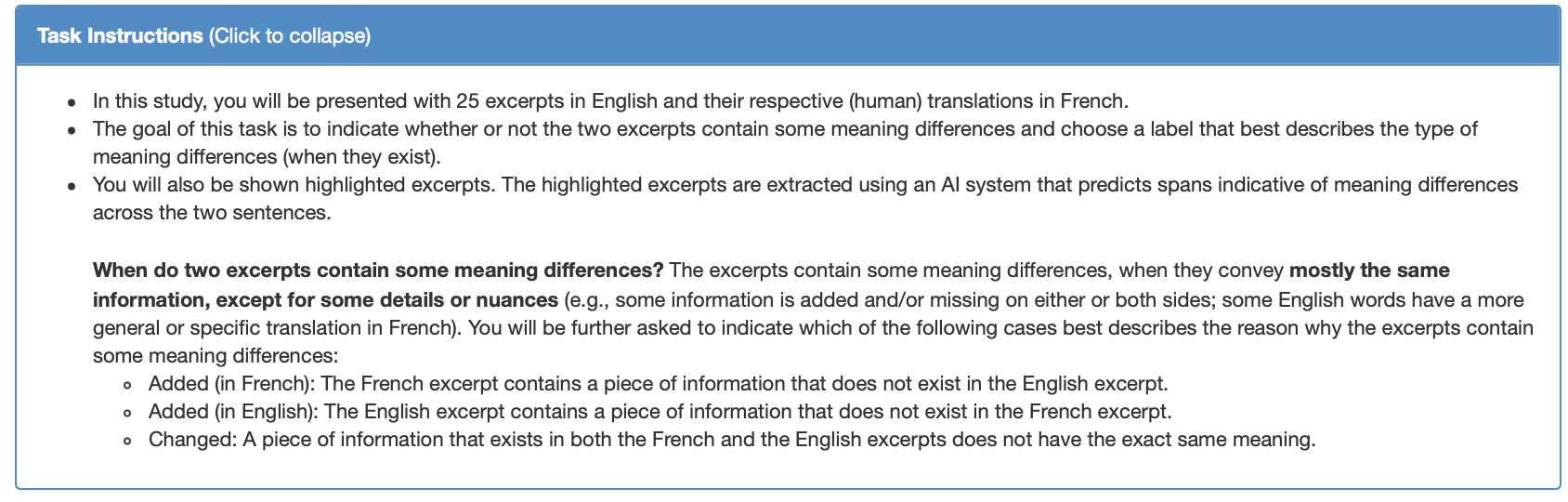}
    \caption{Instructions for application-grounded evaluation I: Annotation of Semantic Divergences.}
    \label{fig:instructions_1}
\end{figure*}


\begin{figure*}
%
%
\begin{subfigure}{.5\textwidth}
  \centering
  \includegraphics[width=.8\linewidth]{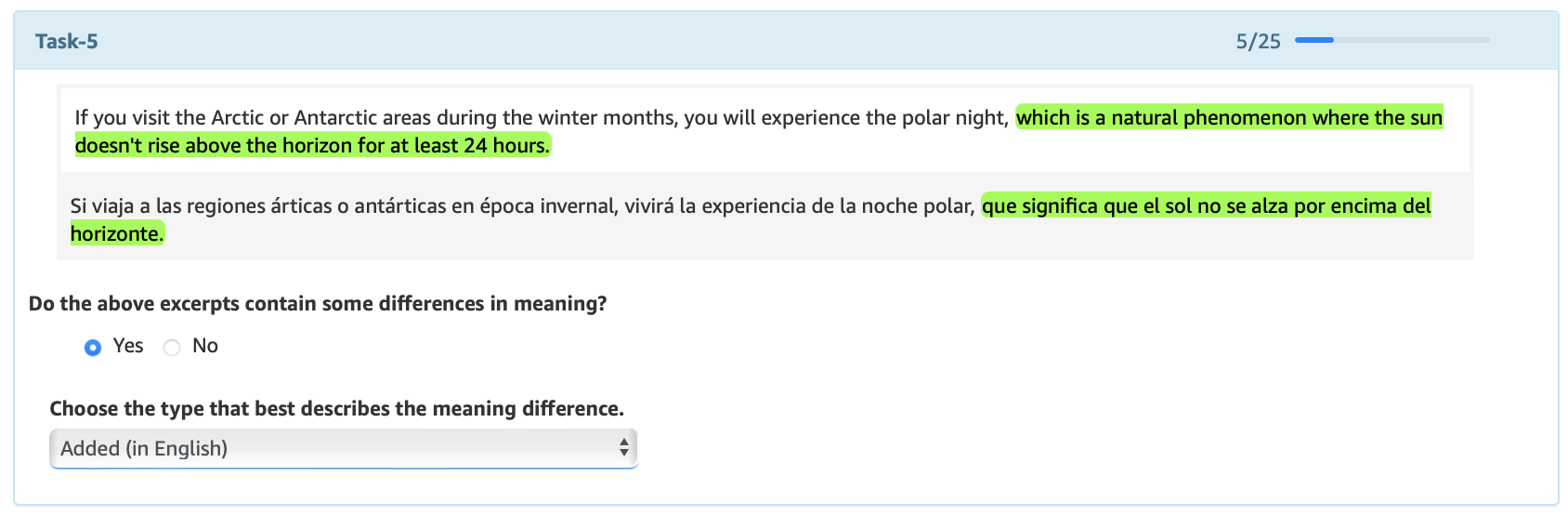}
  \caption{}\label{fig:added_a}
\end{subfigure}
\begin{subfigure}{.5\textwidth}
  \centering
  \includegraphics[width=.8\linewidth]{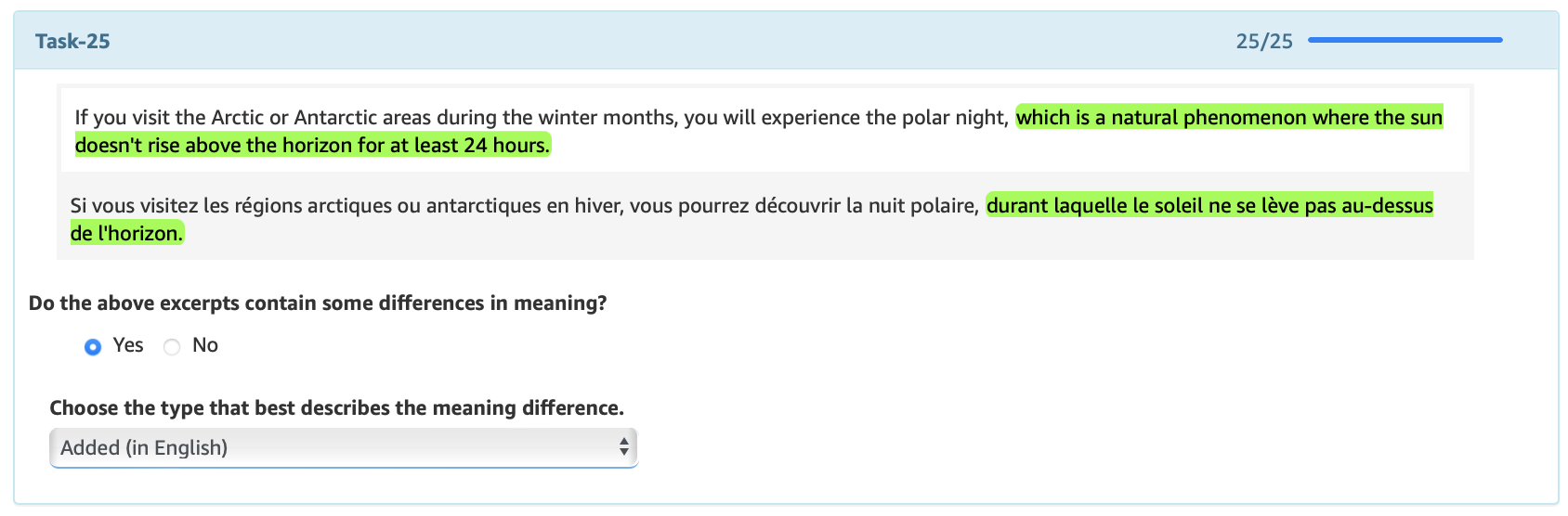}
  \caption{}
  \label{fig:added_b}
\end{subfigure}\vspace*{2mm}
%
%
\begin{subfigure}{.5\textwidth}
  \centering
  \includegraphics[width=.8\linewidth]{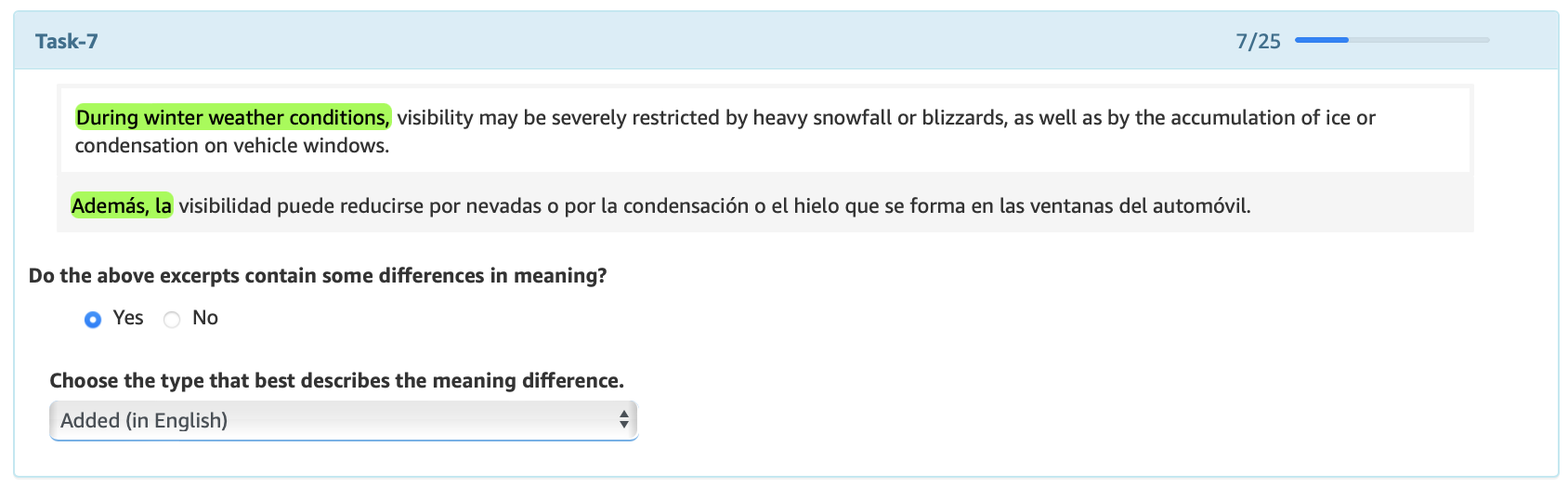}
  \caption{}
  \label{fig:added_c}
\end{subfigure}
\begin{subfigure}{.5\textwidth}
  \centering
  \includegraphics[width=.8\linewidth]{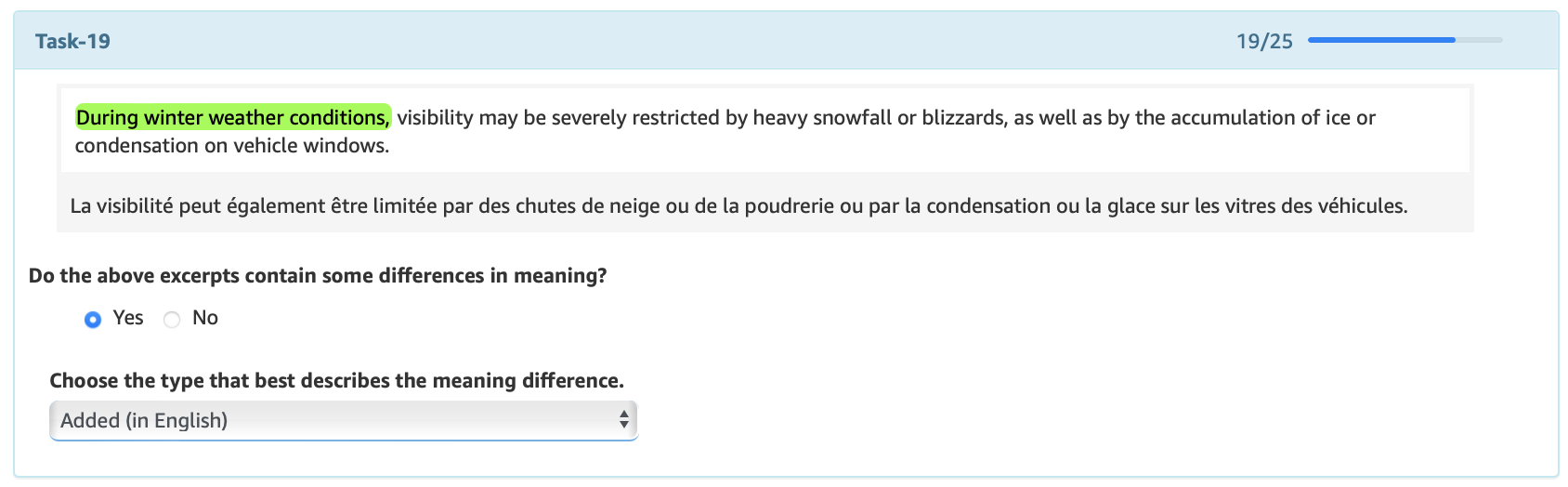}
  \caption{}
  \label{fig:added_d}
\end{subfigure}\vspace*{2mm}
%
%
\begin{subfigure}{.5\textwidth}
  \centering
  \includegraphics[width=.8\linewidth]{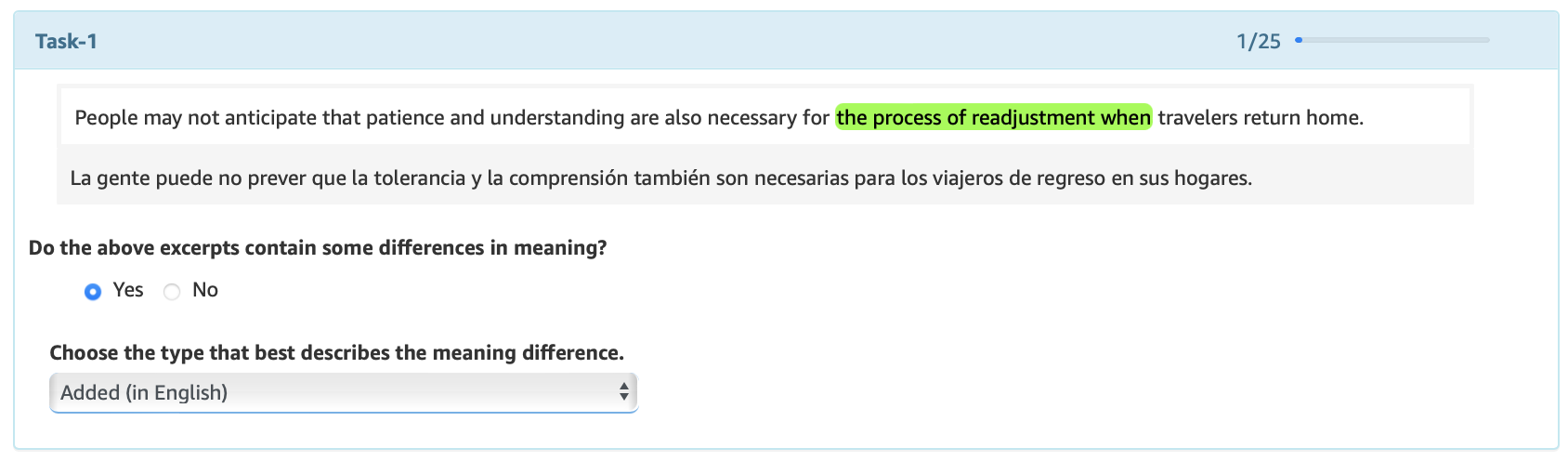}
  \caption{}
  \label{fig:added_e}
\end{subfigure}
\begin{subfigure}{.5\textwidth}
  \centering
  \includegraphics[width=.8\linewidth]{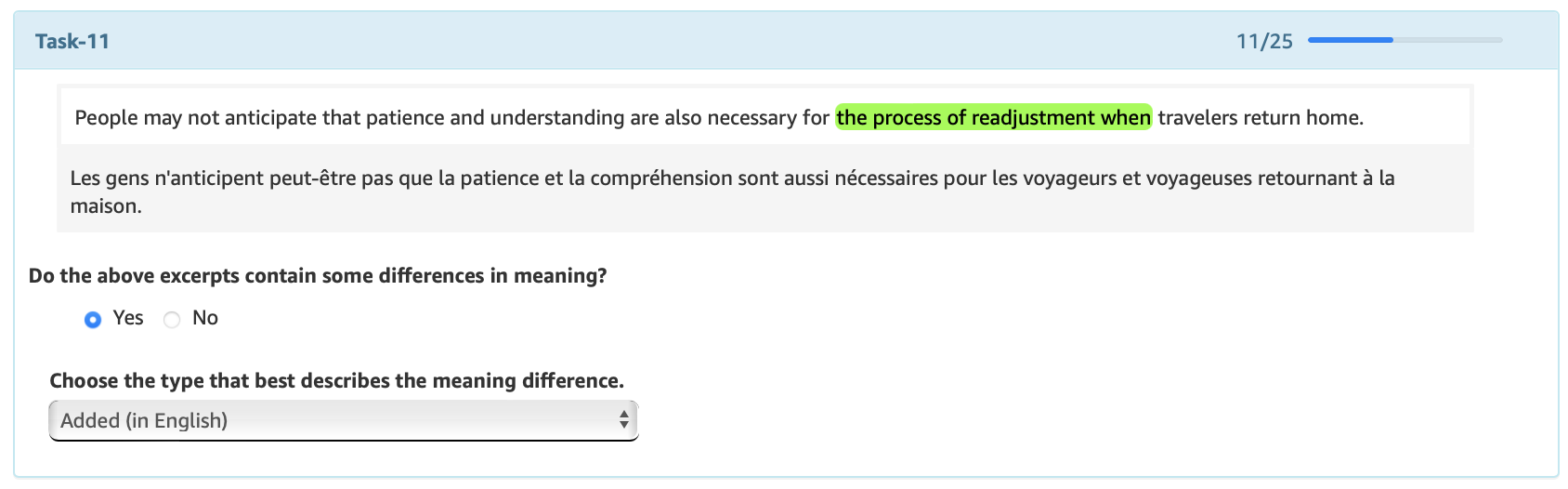}
  \caption{}
  \label{fig:added_f}
\end{subfigure}\vspace*{2mm}
%
%
\begin{subfigure}{.5\textwidth}
  \centering
  \includegraphics[width=.8\linewidth]{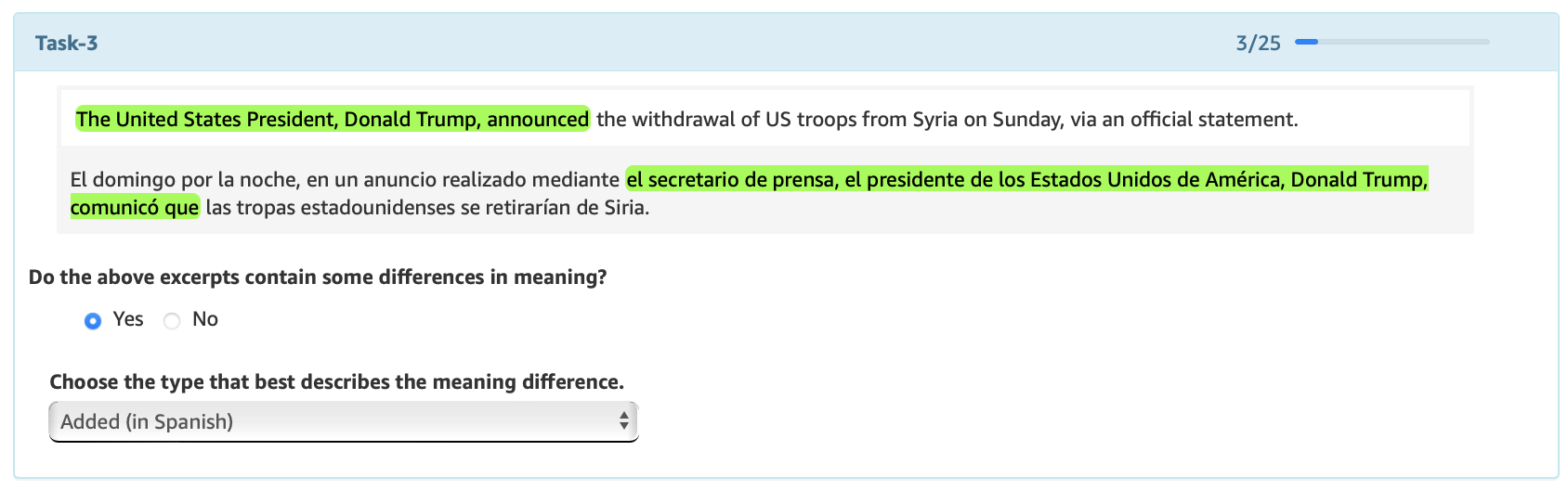}
  \caption{}
  \label{fig:added_g}
\end{subfigure}
\begin{subfigure}{.5\textwidth}
  \centering
  \includegraphics[width=.8\linewidth]{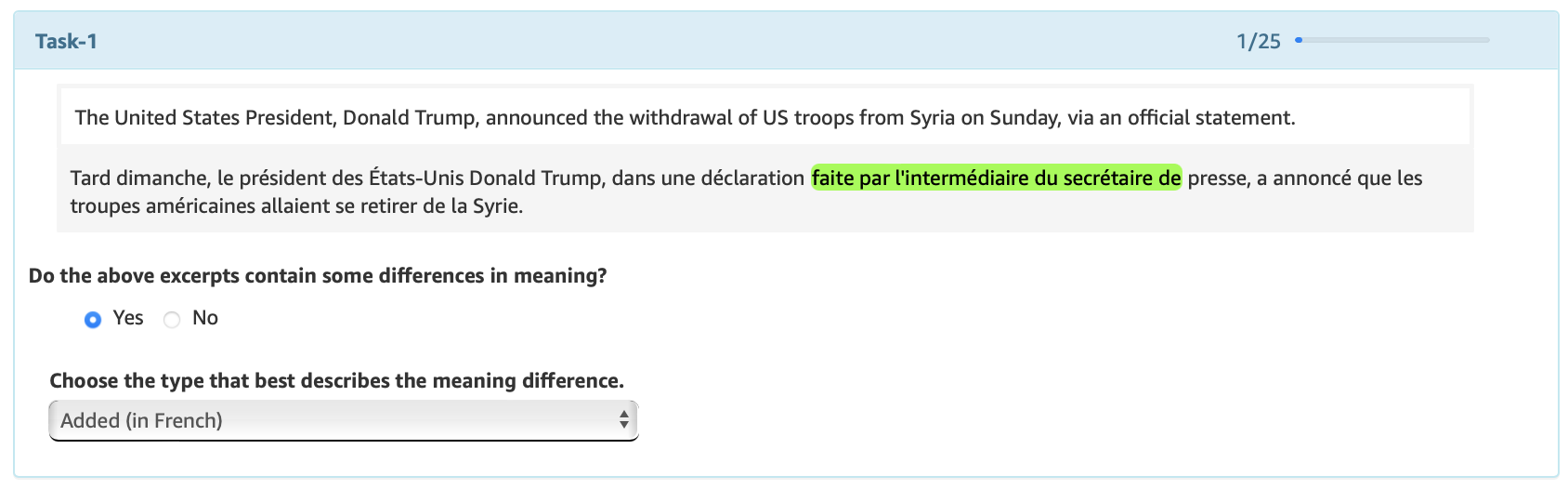}
  \caption{}
  \label{fig:added_h}
\end{subfigure}\vspace*{2mm}
%
\caption{Annotations of fine-grained semantic divergences (English-Spanish on the left and English-French on the righ) reflective of \underline{\textbf{\textit{explicitation}}} and \underline{\textbf{\textit{reduction}}} translation processes. Contrastive highlights subsume added content that is present in one language but missing from the other.}
\label{fig:interface_examples_added_1}
\end{figure*}


\begin{figure*}
%
%
\begin{subfigure}{.5\textwidth}
  \centering
  \includegraphics[width=.8\linewidth]{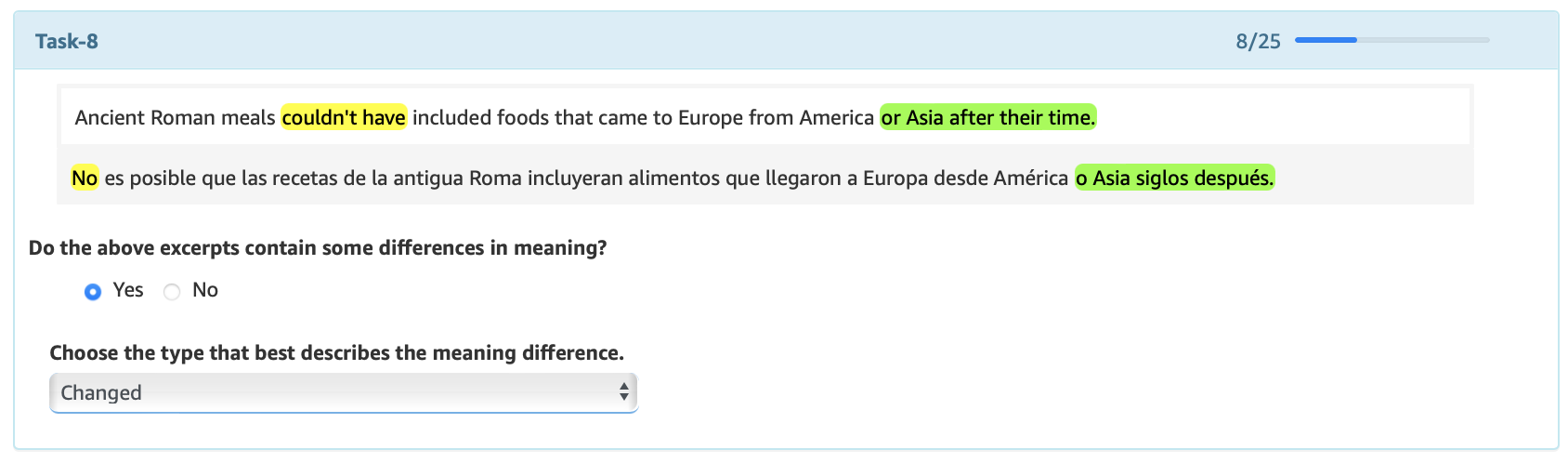}
  \caption{}
  \label{fig:changed_a}
\end{subfigure}
\begin{subfigure}{.5\textwidth}
  \centering
  \includegraphics[width=.8\linewidth]{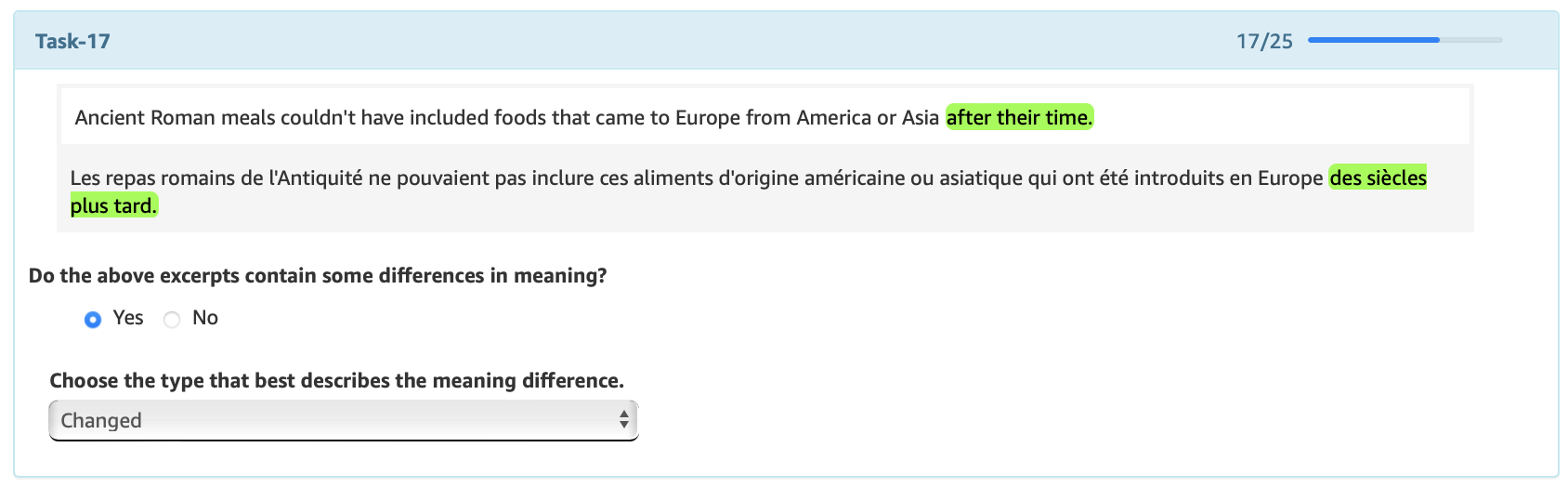}
  \caption{}
  \label{fig:changed_b}
\end{subfigure}\vspace*{2mm}
%
%
\begin{subfigure}{.5\textwidth}
  \centering
  \includegraphics[width=.8\linewidth]{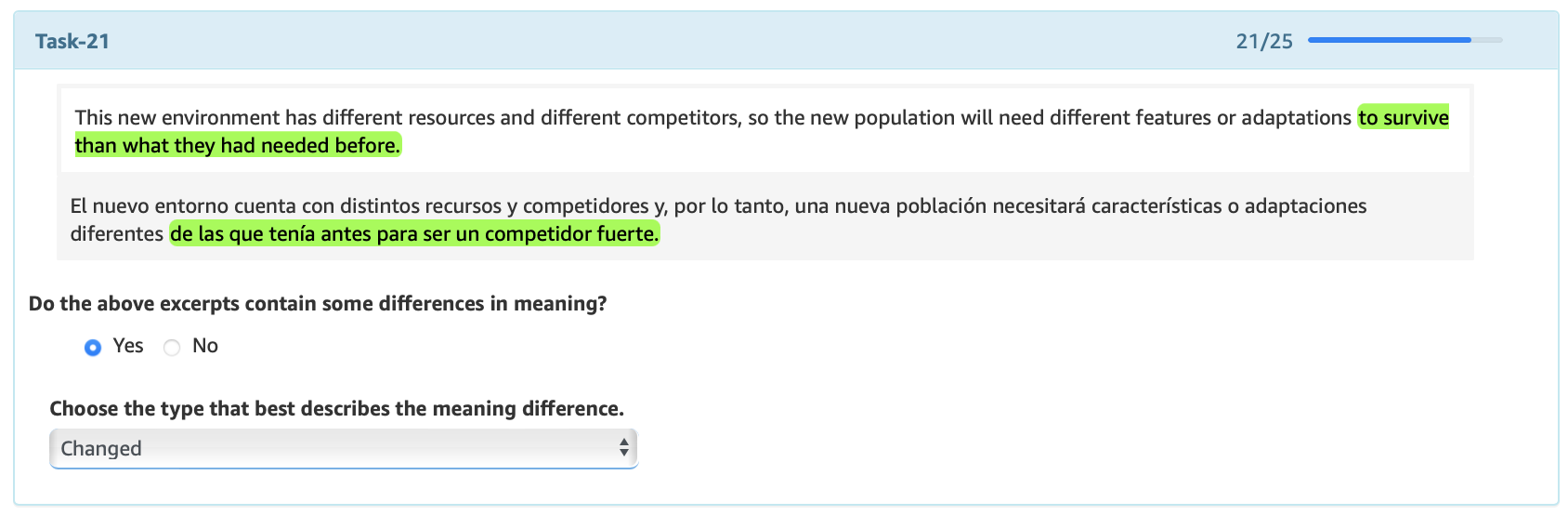}
  \caption{}
  \label{fig:changed_c}
\end{subfigure}
\begin{subfigure}{.5\textwidth}
  \centering
  \includegraphics[width=.8\linewidth]{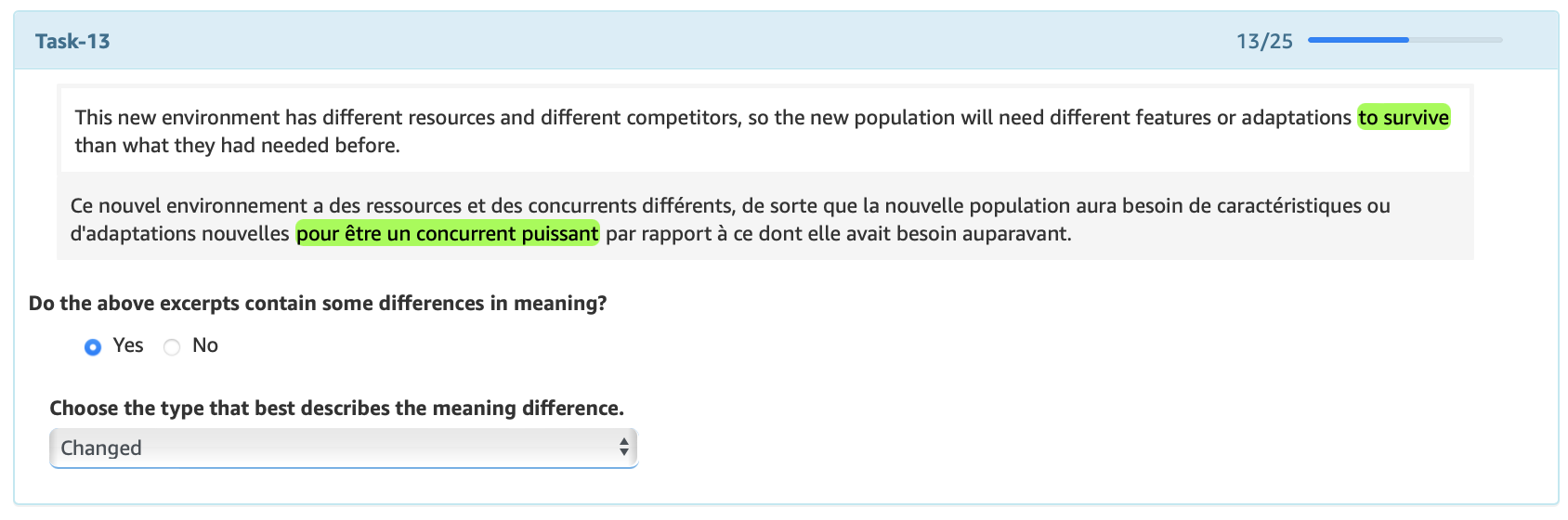}
  \caption{}
  \label{fig:changed_d}
\end{subfigure}\vspace*{2mm}
\caption{Annotations of fine-grained semantic divergences (English-Spanish on the left and English-French on the right) reflective of \underline{\textbf{\textit{modulation}}} translation processes. Contrastive highlights frequently subsume content that does not convey the exact same meaning across languages.}
\label{fig:interface_examples_changed_1}
\end{figure*}


\begin{figure*}
\begin{subfigure}{.5\textwidth}
  \centering
  \includegraphics[width=.8\linewidth]{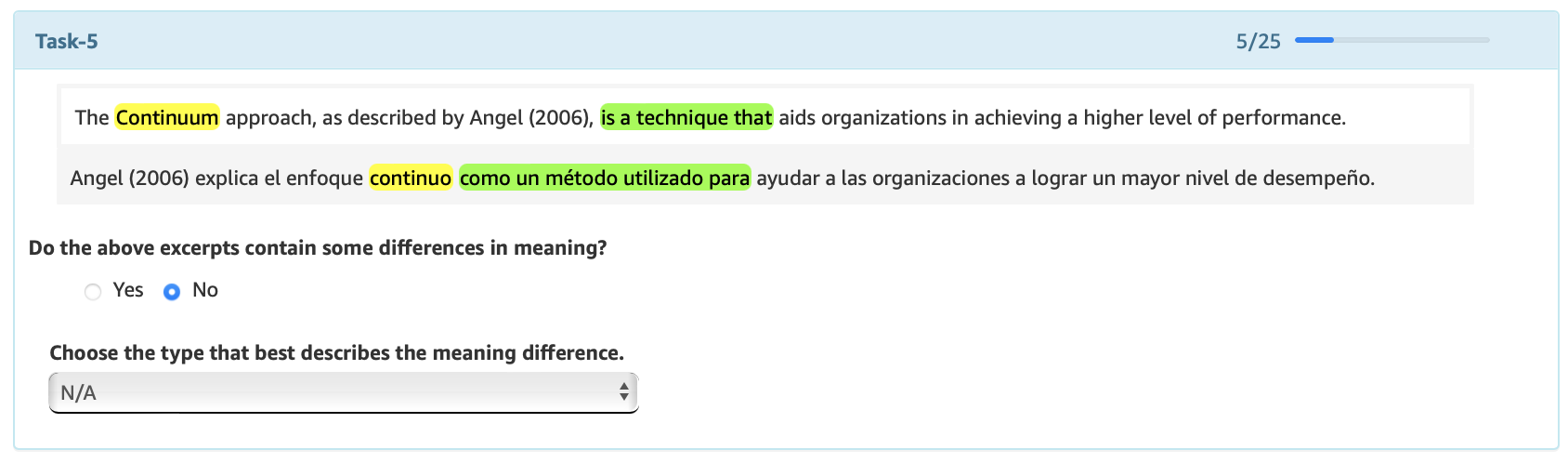
  }
  \caption{}
  \label{fig:para_a}
\end{subfigure}
\begin{subfigure}{.5\textwidth}
  \centering
  \includegraphics[width=.8\linewidth]{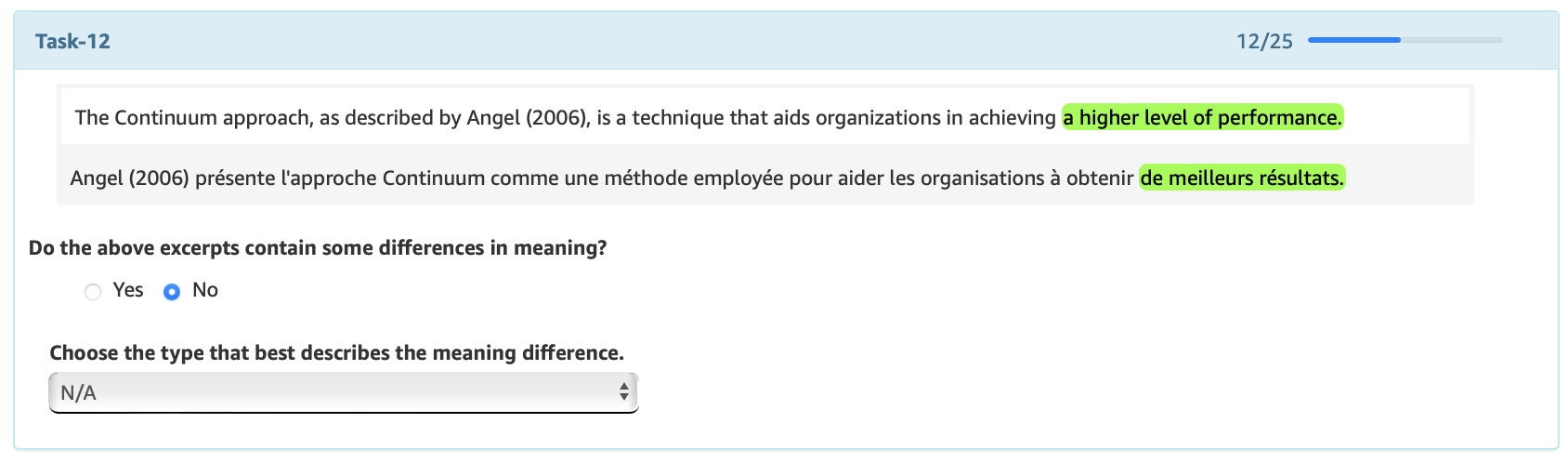}
  \caption{}
  \label{fig:para_b}
\end{subfigure}\vspace*{2mm}
%
\begin{subfigure}{.5\textwidth}
  \centering
  \includegraphics[width=.8\linewidth]{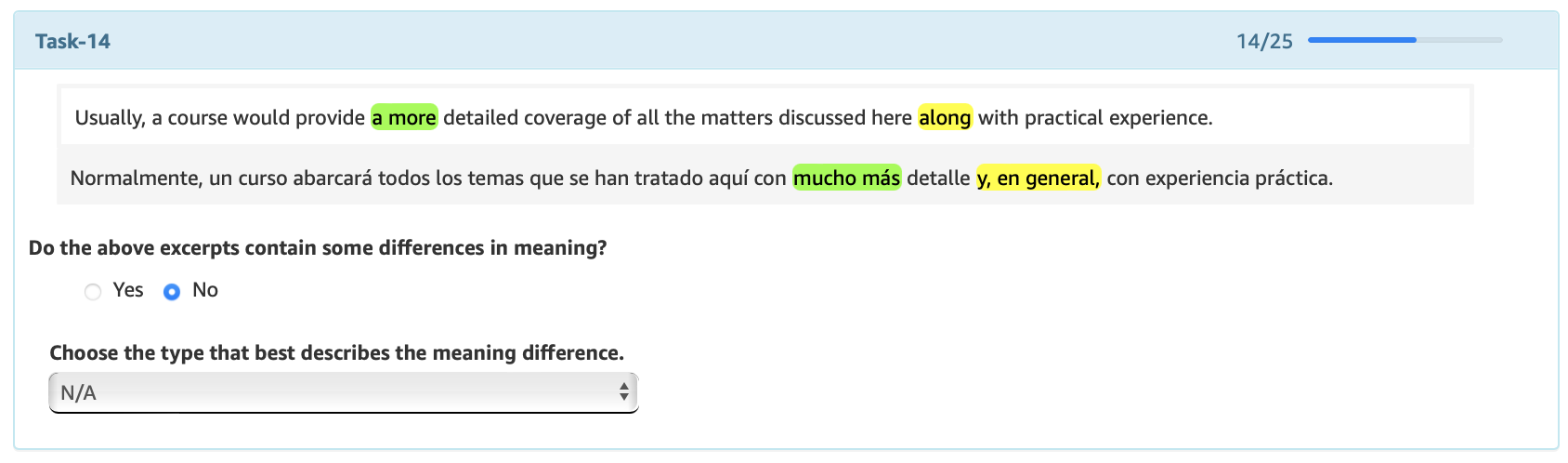}
  \caption{}
  \label{fig:para_c}
\end{subfigure}
\begin{subfigure}{.5\textwidth}
  \centering
  \includegraphics[width=.8\linewidth]{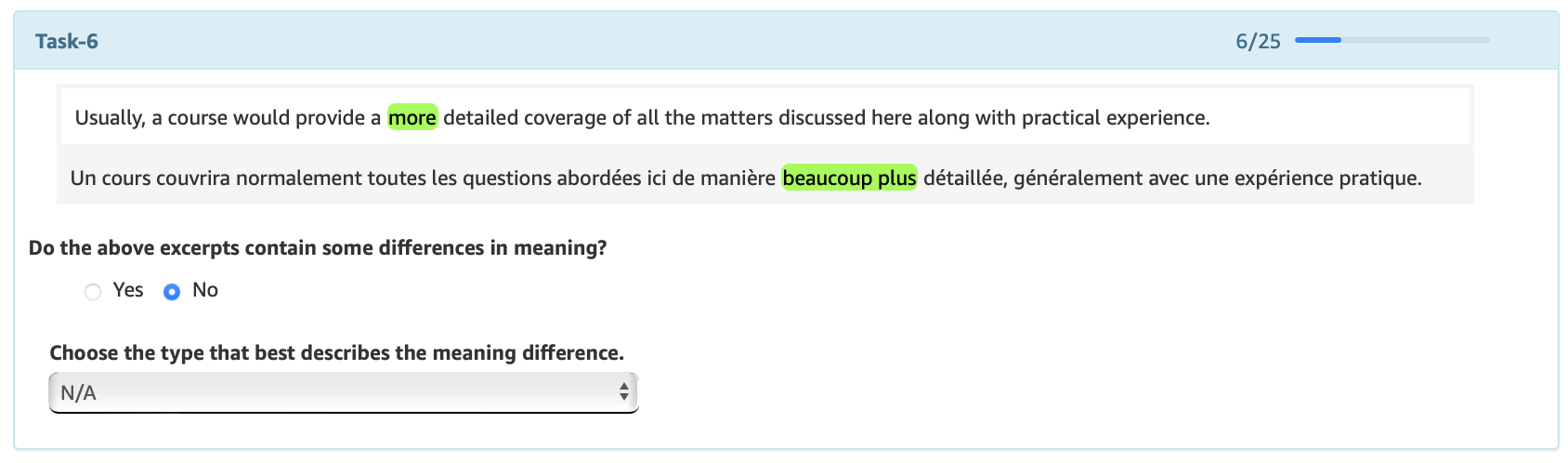}
  \caption{}
  \label{fig:para_d}
\end{subfigure}\vspace*{2mm}
%
\begin{subfigure}{.5\textwidth}
  \centering
  \includegraphics[width=.8\linewidth]{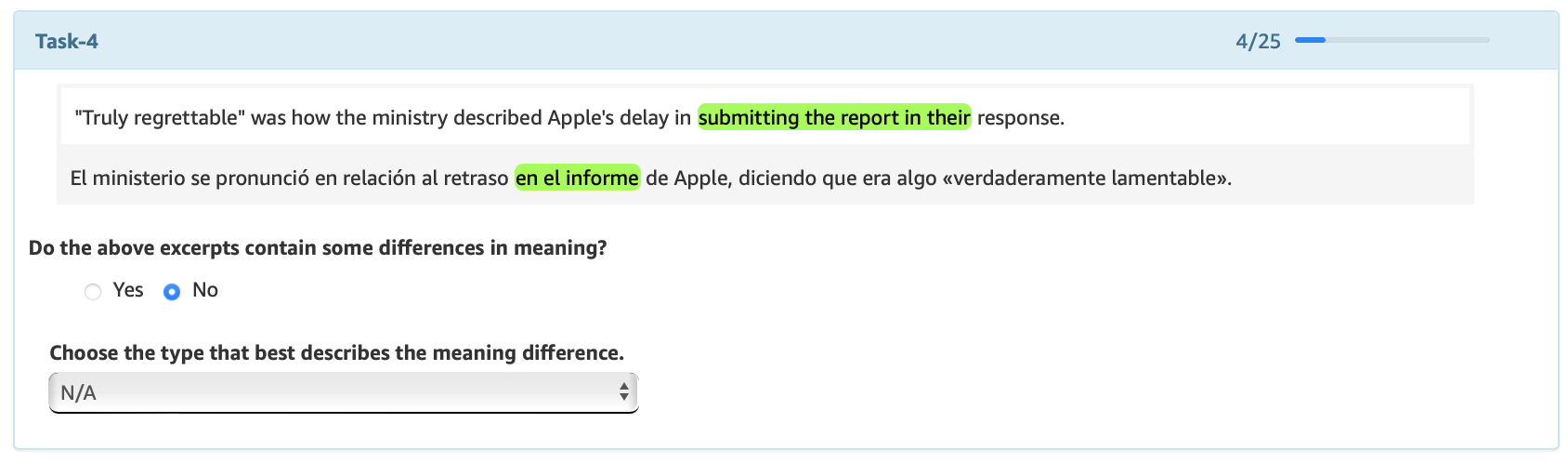}
  \caption{}
  \label{fig:para_e}
\end{subfigure}
\begin{subfigure}{.5\textwidth}
  \centering
  \includegraphics[width=.8\linewidth]{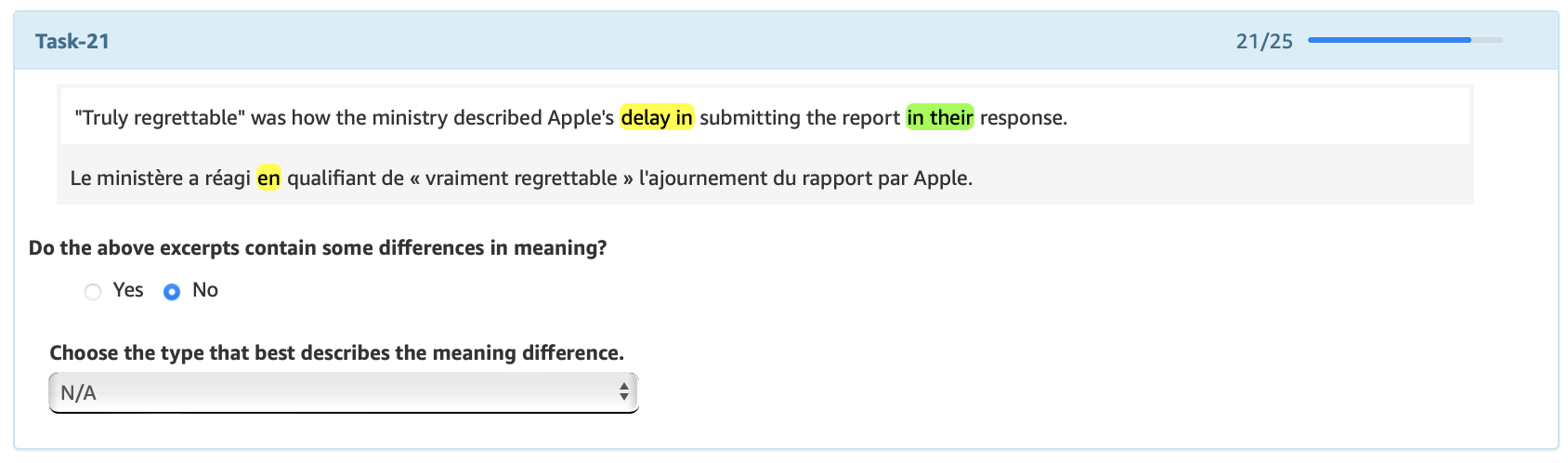}
  \caption{}
  \label{fig:para_f}
\end{subfigure}\vspace*{2mm}
%
\begin{subfigure}{.5\textwidth}
  \centering
  \includegraphics[width=.8\linewidth]{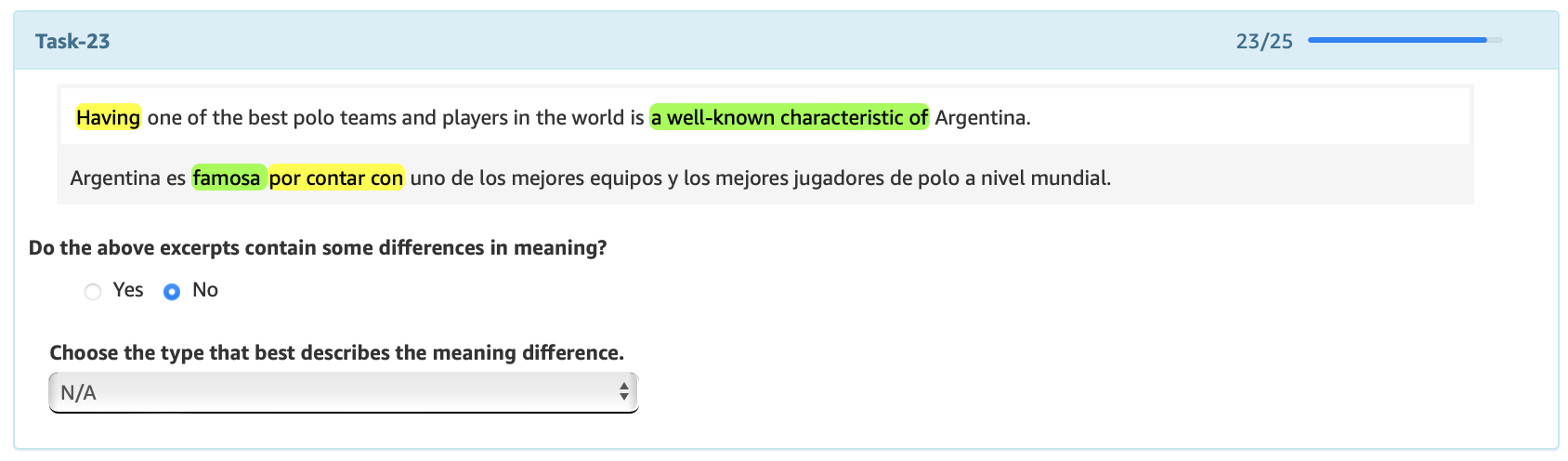}
  \caption{}
  \label{fig:para_g}
\end{subfigure}
\begin{subfigure}{.5\textwidth}
  \centering
  \includegraphics[width=.8\linewidth]{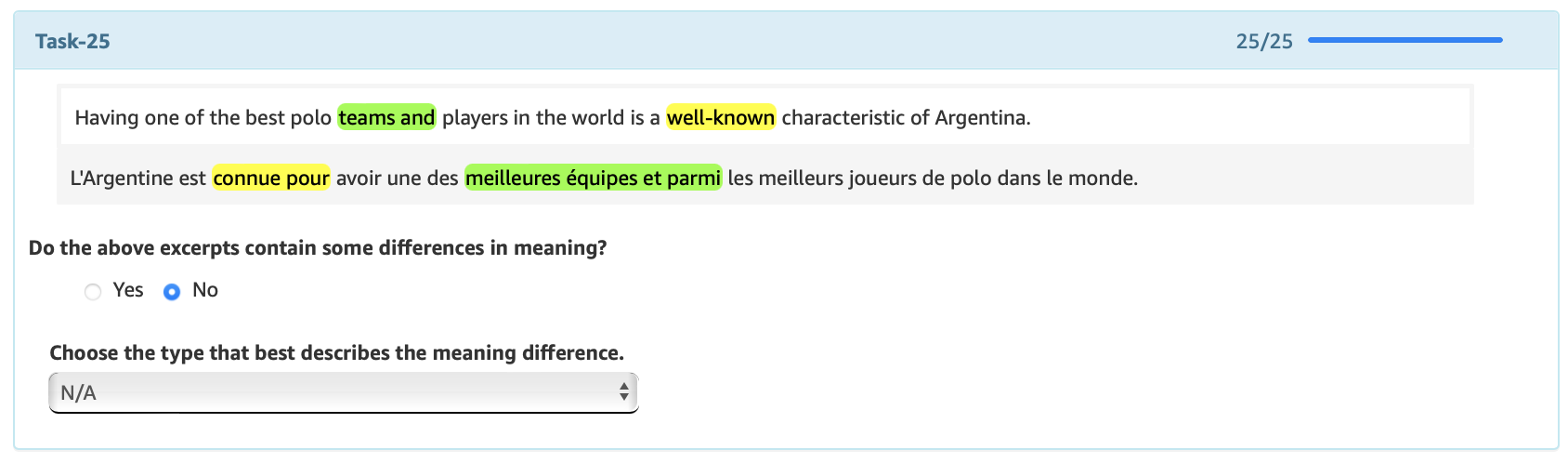}
  \caption{}
  \label{fig:para_h}
\end{subfigure}\vspace*{2mm}
\caption{Annotations of semantically equivalent pairs (English-Spanish on the left and English-French on the right) that reflect syntactic divergences. Contrastive highlights surface \underline{\textbf{\textit{false positive}}} segments.}
\label{fig:interface_examples_paraphrased_1}
\end{figure*}

\clearpage

\begin{figure*}[h]
    \centering
    \includegraphics[width=1.\linewidth]{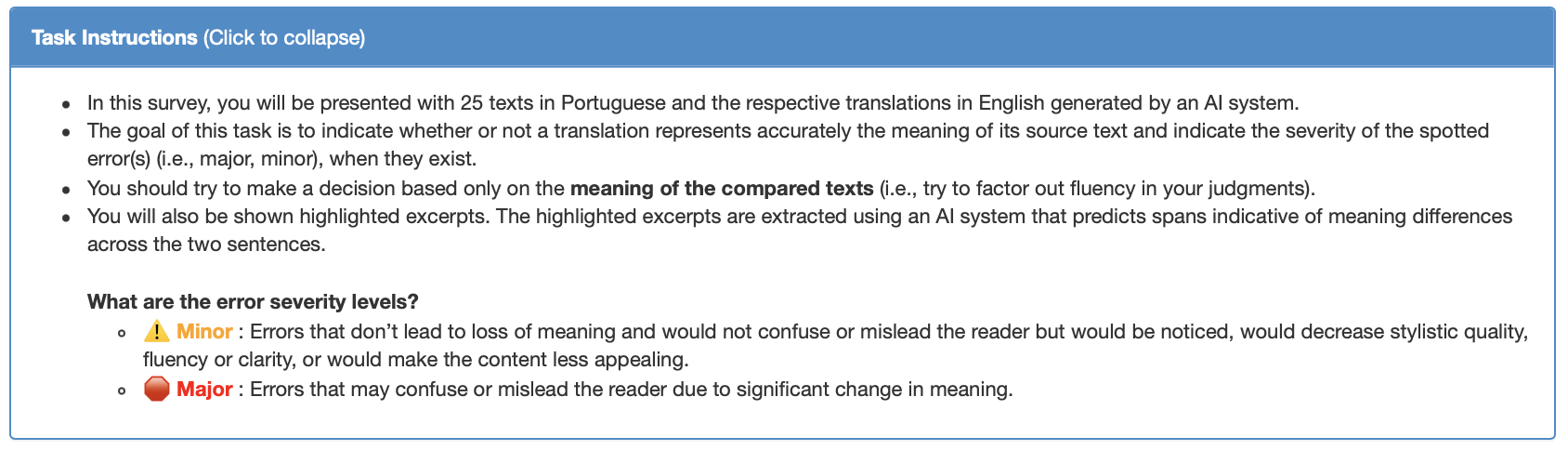}
    \caption{Instructions for application-grounded evaluation II: Critical Error Detection.}
    \label{fig:instructions_2}
\end{figure*}

\begin{figure*}
%
\begin{subfigure}{.5\textwidth}
  \centering
  \includegraphics[width=.8\linewidth]{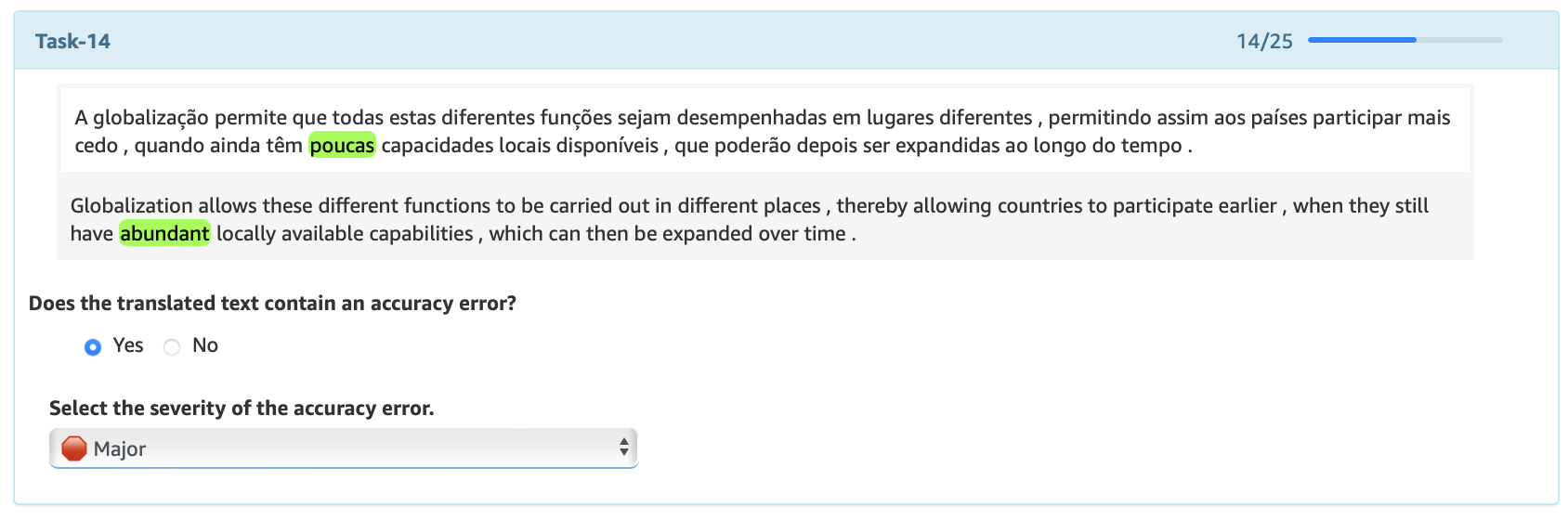}
  \caption{\textbf{Major error} (negation)}
  \label{fig:ced_negation_1}
\end{subfigure}
%
\begin{subfigure}{.5\textwidth}
  \centering
  \includegraphics[width=.8\linewidth]{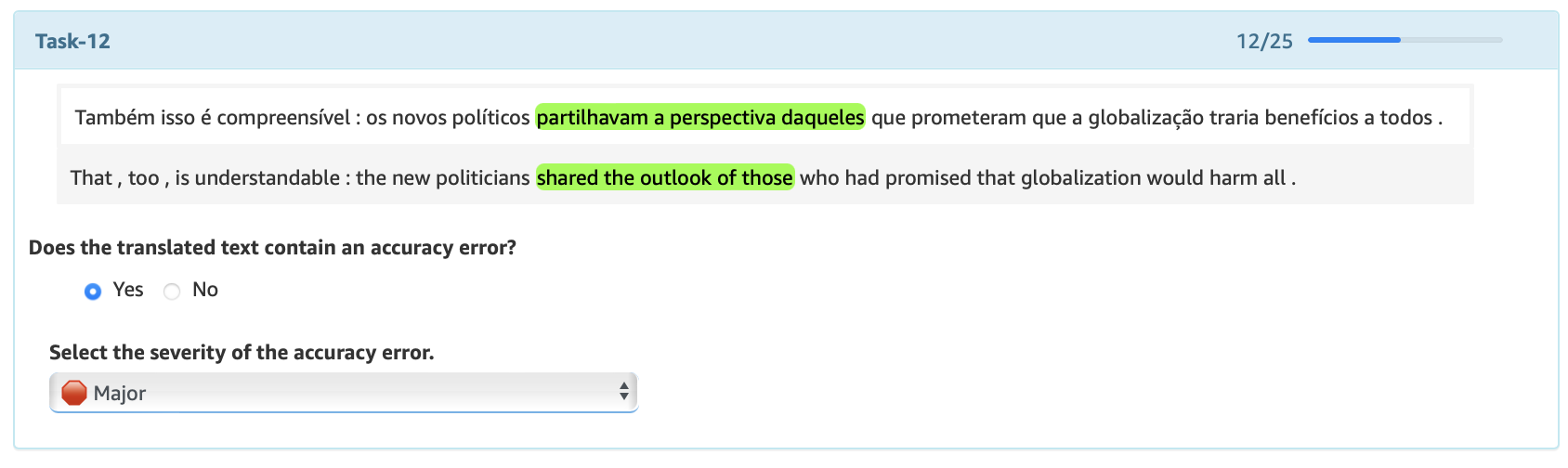}
  \caption{\textbf{Major error} (negation)}
  \label{fig:ced_negation_2}
\end{subfigure}\vspace*{2mm}
%
\begin{subfigure}{.5\textwidth}
  \centering
  \includegraphics[width=.8\linewidth]{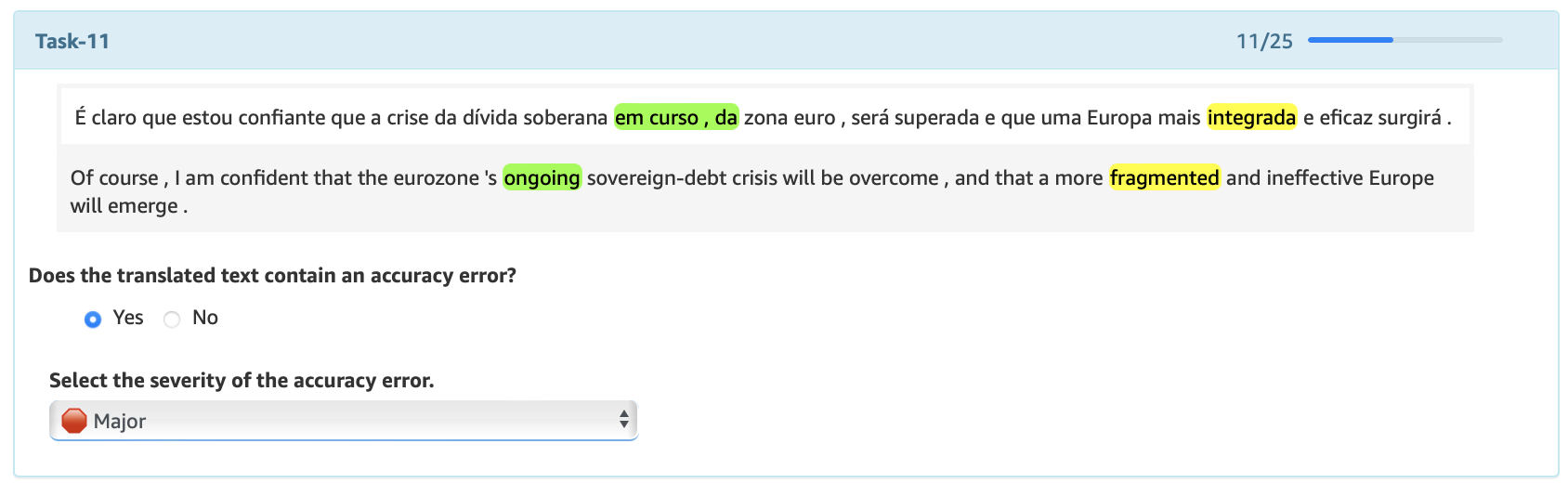}
  \caption{\textbf{Major error} (negation)}
  \label{fig:ced_negation_3}
\end{subfigure}
\begin{subfigure}{.5\textwidth}
  \centering
  \includegraphics[width=.8\linewidth]{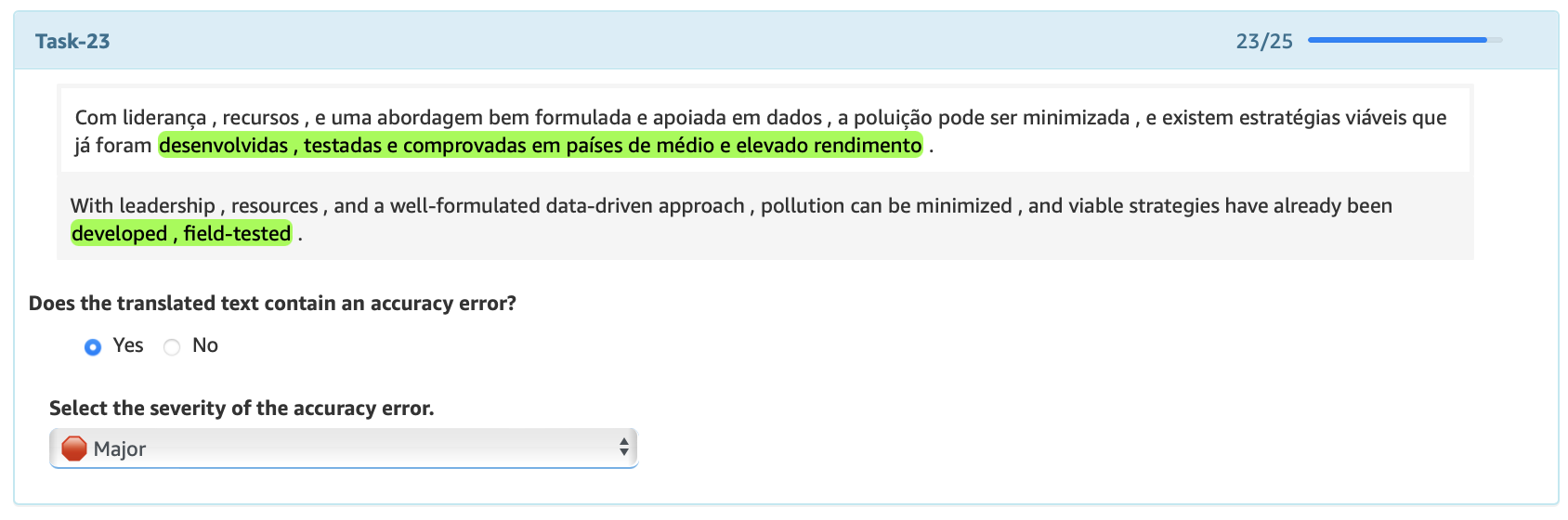}
  \caption{\textbf{Major error} (hallucination)}
  \label{fig:ced_wmt_1}
\end{subfigure}\vspace*{2mm}
%
\begin{subfigure}{.5\textwidth}
  \centering
  \includegraphics[width=.8\linewidth]{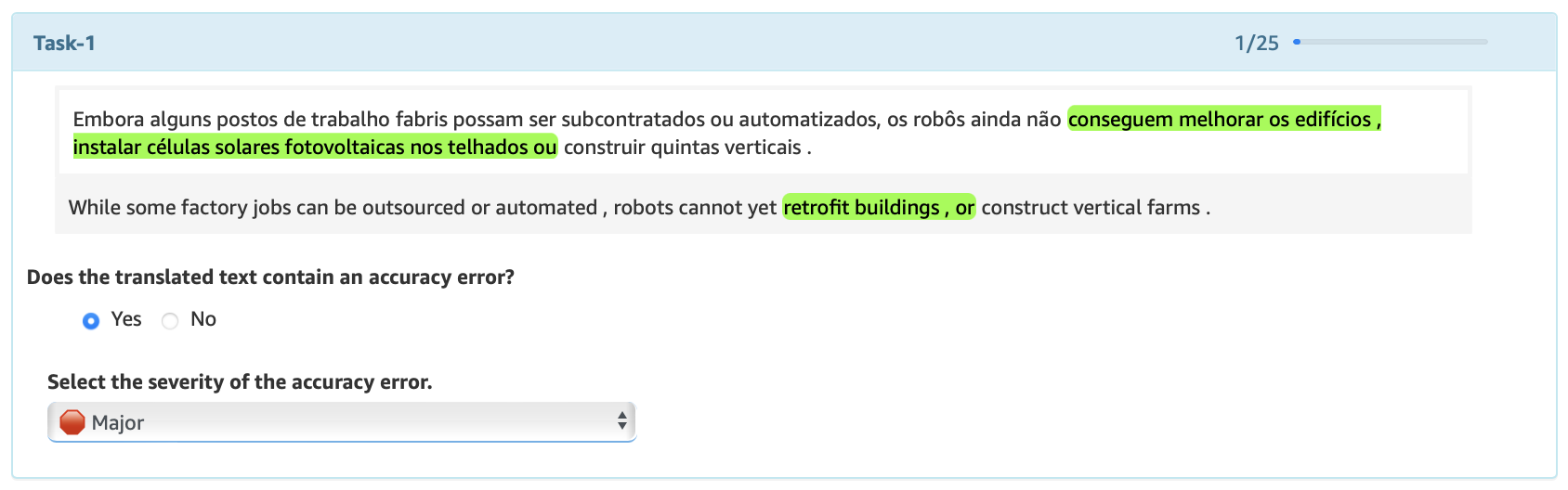}
  \caption{\textbf{Major error} (hallucination)}
  \label{fig:ced_wmt_2}
\end{subfigure}
\begin{subfigure}{.5\textwidth}
  \centering
  \includegraphics[width=.8\linewidth]{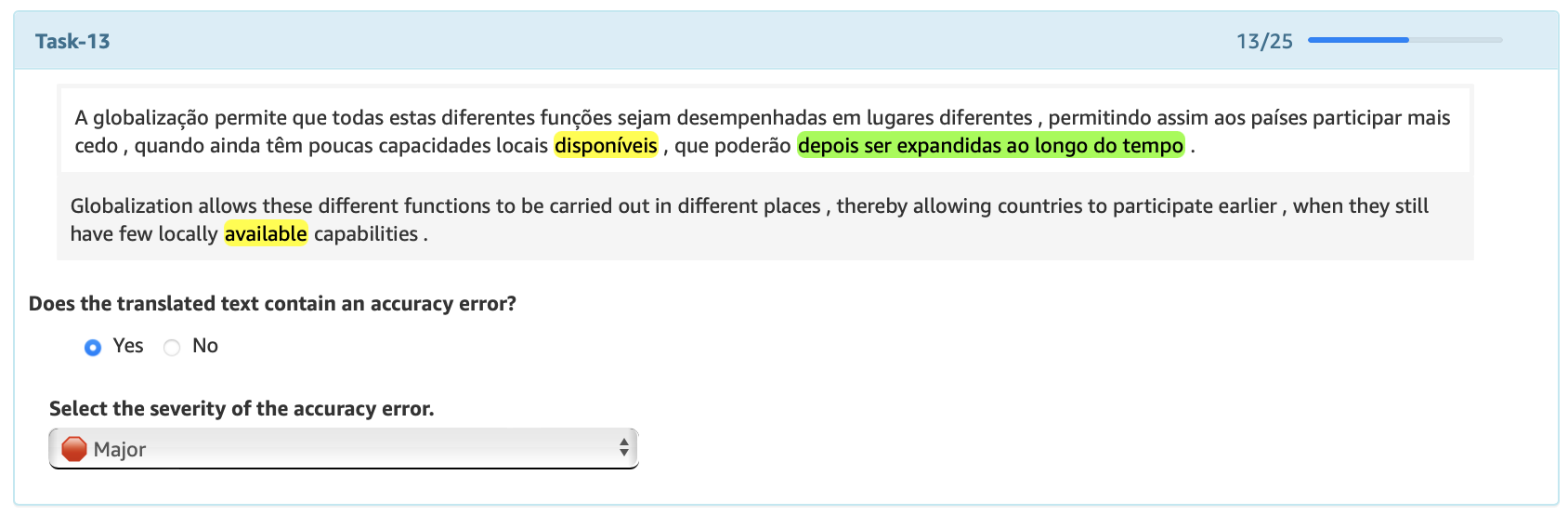}
  \caption{\textbf{Major error} (hallucination)}
  \label{fig:ced_wmt_3}
\end{subfigure}\vspace*{2mm}
%
\begin{subfigure}{.5\textwidth}
  \centering
  \includegraphics[width=.8\linewidth]{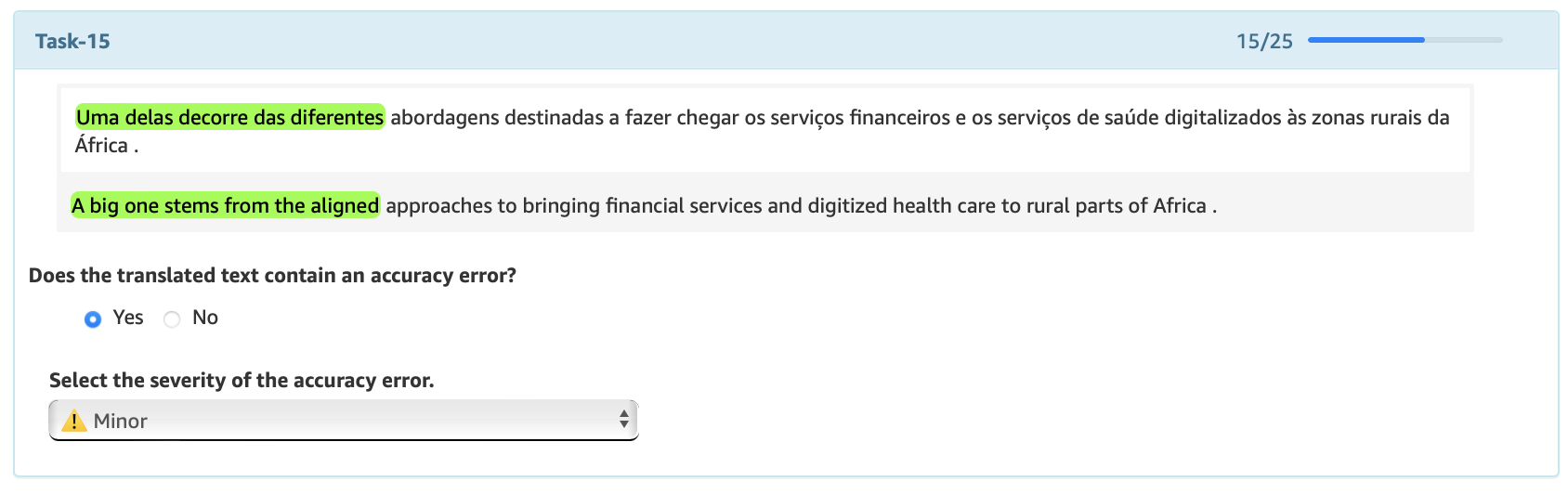}
  \caption{\textbf{Minor error}}
  \label{fig:ced_minor_1}
\end{subfigure}
\begin{subfigure}{.5\textwidth}
  \centering
  \includegraphics[width=.8\linewidth]{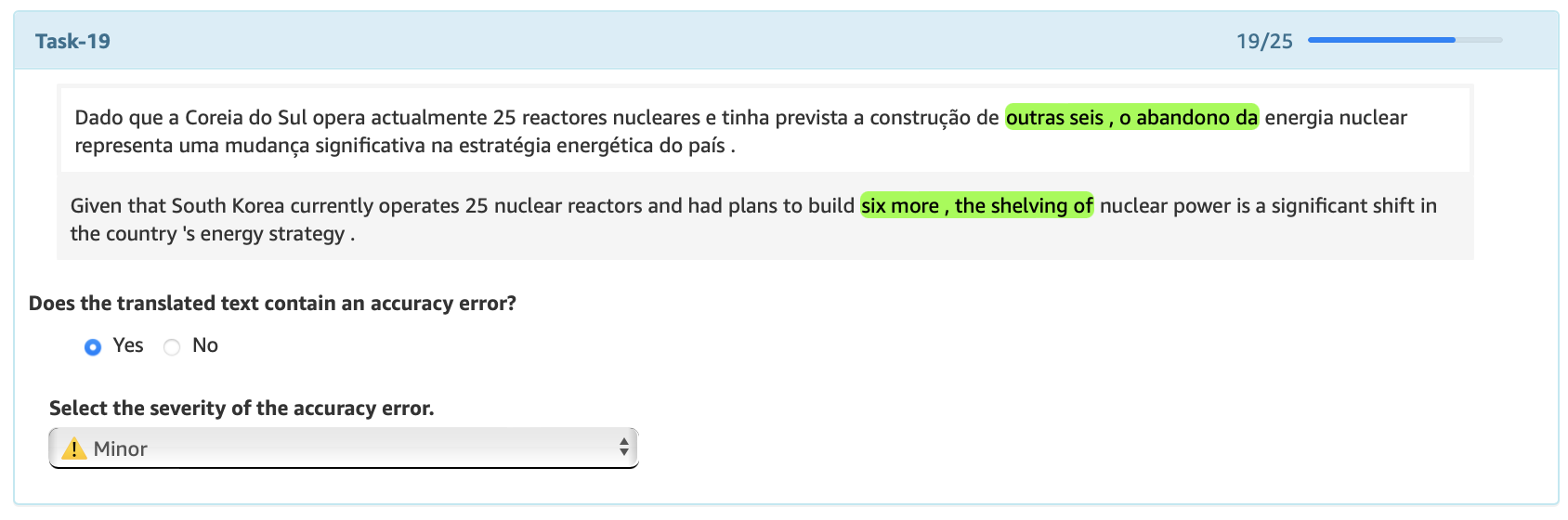}
  \caption{\textbf{Minor error}}
  \label{fig:ced_minor_2}
\end{subfigure}\vspace*{2mm}
%
\begin{subfigure}{.5\textwidth}
  \centering
  \includegraphics[width=.8\linewidth]{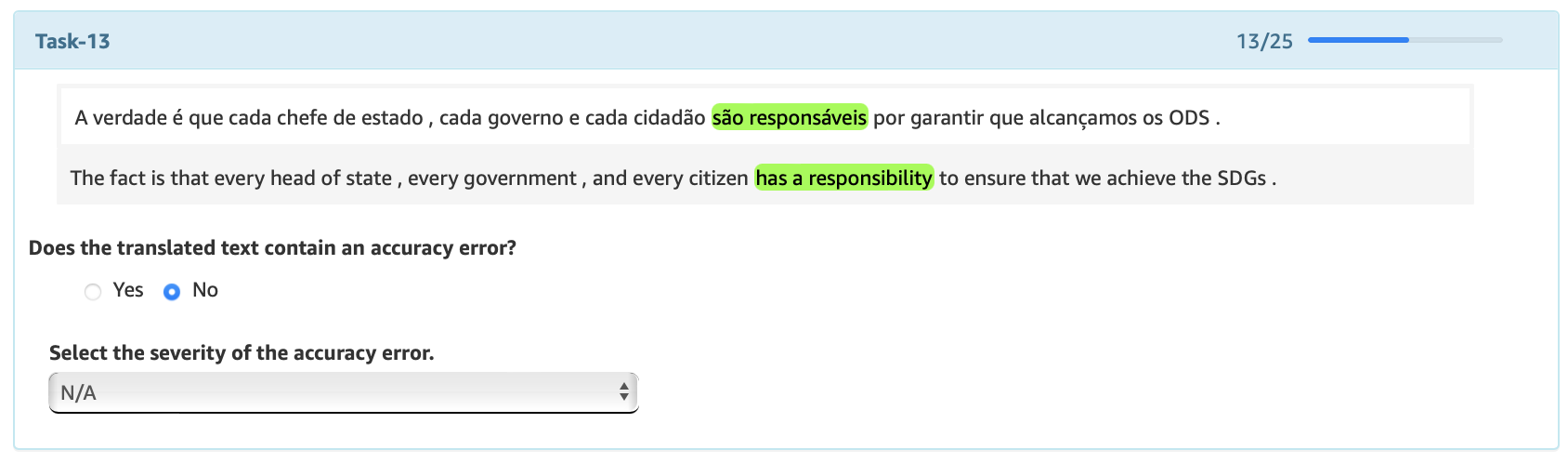}
  \caption{\textbf{No error}}
  \label{fig:ced_na_1}
\end{subfigure}
\begin{subfigure}{.5\textwidth}
  \centering
  \includegraphics[width=.8\linewidth]{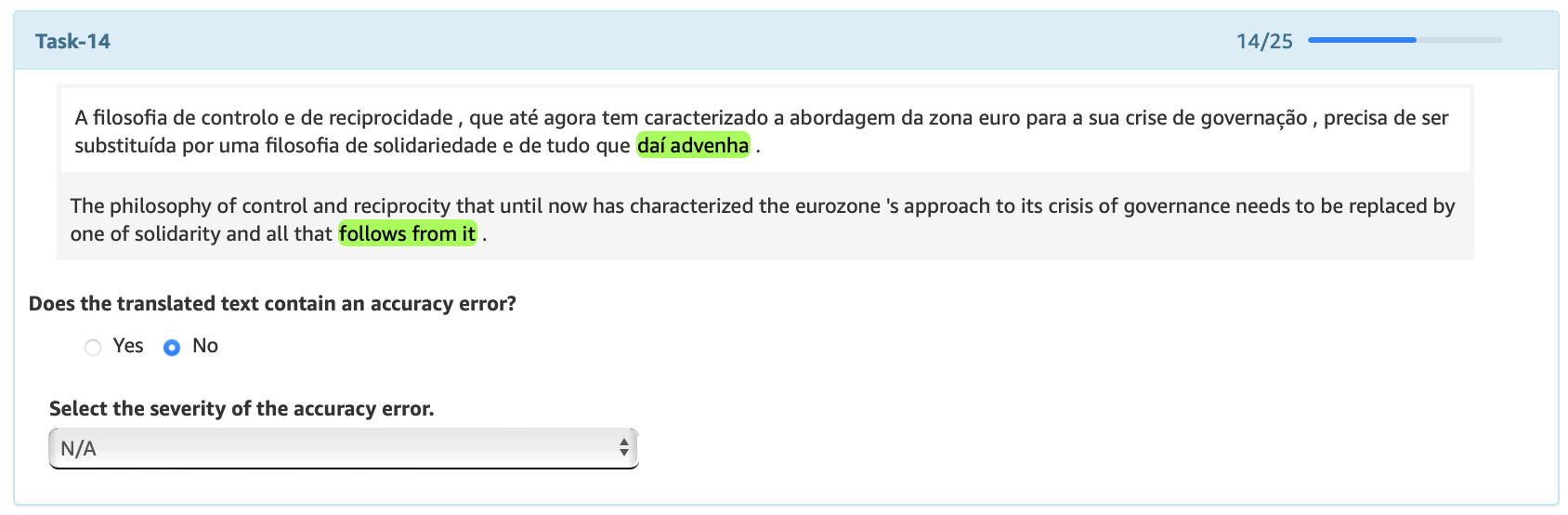}
  \caption{\textbf{No error}}
  \label{fig:ced_na_2}
\end{subfigure}
\caption{
Annotations of accuracy errors in Portuguese translations of English texts. Contrastive highlights frequently surface meaning differences across the compared texts.
}\label{fig:interface_examples_2}
\end{figure*}
\label{sec:appendix}

\end{document}